\documentclass[10pt,twocolumn,letterpaper]{article}

\usepackage{3dv}
\usepackage{times}
\usepackage{epsfig}
\usepackage{graphicx}
\usepackage{amsmath}
\usepackage{amssymb}
\usepackage{authblk}
\usepackage{subcaption}
\usepackage{appendix}
\usepackage{titling}
    \pretitle{\vspace*{-3pt}\begin{center}\Large \bf}
    \posttitle{\vspace*{24pt}\par\end{center}}
    \preauthor{\large\begin{center}\par}
    \postauthor{\vspace*{12pt}\par\end{center}}
    \predate{}
    \date{}
    \postdate{}
\usepackage[pagebackref=true,breaklinks=true,colorlinks,bookmarks=false]{hyperref}
\usepackage{multirow}
\usepackage{multicol}
\usepackage{subcaption}
\usepackage{pifont}% http://ctan.org/pkg/pifont
\newcommand{\cmark}{\ding{51}}%
\newcommand{\xmark}{\ding{55}}%
\threedvfinalcopy % *** Uncomment this line for the final submission

\def\threedvPaperID{349} % *** Enter the 3DV Paper ID here
\def\httilde{\mbox{\tt\raisebox{-.5ex}{\symbol{126}}}}

\ifthreedvfinal\fi
\begin{document}

%%%%%%%%% TITLE
\title{PolyNet: Polynomial Neural Network for 3D Shape Recognition with PolyShape Representation}

\author{
Mohsen Yavartanoo$^{1}$\quad Shih-Hsuan Hung$^{2}$\quad Reyhaneh Neshatavar$^{1}$\quad Yue Zhang$^{2}$\quad Kyoung Mu Lee$^{1}$\\
\hspace{1.5cm}$^{1}$SNU ECE \& ASRI \hspace{4.75cm}
$^{2}$Oregon State University\\{\small \texttt {\{myavartanoo,reyhanehneshat,kyoungmu\}@snu.ac.kr} \hspace{1.0cm} \texttt{\{hungsh,zhangyue\}@oregonstate.edu}}
}

\maketitle
\thispagestyle{empty}
\ifthreedvfinal\thispagestyle{empty}\fi

%%%%%%%%% ABSTRACT
\begin{abstract}
3D shape representation and its processing have substantial effects on 3D shape recognition. The polygon mesh as a 3D shape representation has many advantages in computer graphics and geometry processing. However, there are still some challenges for the existing deep neural network (DNN)-based methods on polygon mesh representation, such as handling the variations in the degree and permutations of the vertices and their pairwise distances. To overcome these challenges, 
we propose a DNN-based method (PolyNet) and a specific polygon mesh representation (PolyShape) with a multi-resolution structure.
PolyNet contains two operations; (1) a polynomial convolution (PolyConv) operation with learnable coefficients, which learns continuous distributions as the convolutional filters to share the weights across different vertices, and (2) a polygonal pooling (PolyPool) procedure by utilizing the multi-resolution structure of PolyShape to aggregate the features in a much lower dimension. 
Our experiments demonstrate the strength and the advantages of PolyNet on both 3D shape classification and retrieval tasks compared to existing polygon mesh-based methods and its superiority in classifying graph representations of images. The code is publicly available from this  \href{https://myavartanoo.github.io/polynet/}{link}.
\end{abstract}

%%%%%%%%% BODY TEXT
\section{Introduction}
\label{sec:intro}

%%%%%%%%%%%%%%%%%%%%%%%%%%%%%%%%%%%%%%%%%%%%%%%%%%%%%%%
In recent years, increasing applications of 3D shapes representation have made it a fundamental problem in computer vision, computer graphics, and augmented reality. The structure and the high-quality appearance of the representation significantly impact many tasks, such as 3D shape classification and retrieval.
With the advent of deep neural network (DNN) architectures, several methods have been proposed to learn 3D shapes. 
Generally, these methods can be categorized into four groups based on the input shape representation; point clouds~\cite{Qi2016PointNetDL,Qi2017PointNetDH}, voxel grids~\cite{7298801,7353481}, 2D projections~\cite{su15mvcnn,7273863,3dor.20171045,DBLP:journals/corr/abs-1801-10130, Yavartanoo2018SPNetD3}, and polygon meshes~\cite{Hanocka2019MeshCNNAN,Feng2018MeshNetMN,8100059,Masci:2015:GCN:2919341.2920992}. 
The point clouds suffer from harsh noises and wasted substantial structural information of the 3D shapes.
The voxel grids require large memory, and also rendering voxel grids generates unnecessarily voluminous data and quantization artifacts. 
Furthermore, 2D projection representations encounter severe self-occlusions.

By contrast, a polygon mesh is a collection of vertices and faces that defines a 3D shape smoothly and entirely. Therefore, this representation contains structural information without any harsh noises, severe artifacts, and self-occlusions. 
Additionally, it is a memory-efficient representation that can store the full geometry details by reducing unnecessary voluminous data.
However, in polygon mesh-based methods, weight sharing is still a challenging problem due to variations in the degree of vertices, the permutation of adjacent vertices, and their pairwise distances.

To overcome the limitations of the polygon mesh-based methods, in this work, we propose PolyNet, a novel network that can effectively learn and extract features of a polygon mesh representation of 3D shapes by a continuous polynomial convolution (PolyConv).
PolyConv is a polynomial function with learnable coefficients which learns continuous distributions as the convolutional filters to share the corresponding weights among the features of the vertices in the local patches made from each vertex and its adjacent vertices on the surface. 
This operation is invariant to the number of adjacent vertices, their permutations, and their pairwise distances nearby the central vertex in the local patch.
Moreover, we design PolyShape representation, a specific polygon mesh representation with a multi-resolution structure.
We utilize this multi-resolution attribute to design our PolyPool operation and apply it after each PolyConv layer.
This PolyPool operation reduces the mesh resolution by a fixed factor at each layer. 
We achieve the best classification accuracy and mean Average Precision (mAP) compared to the previous methods based on voxel grid and polygon mesh and comparable performance to point cloud-based methods. We also show the superiority of our designed PolyConv on the challenging 75 Superpixel MNIST dataset.
%%%%%%%%%%%%%%%%%%%%%%%%%%%%%%%%%%%%%%%%%%%%%%%%%%%%%%%
We summarize the main {\bf contributions} of our method as follows:         
\begin{itemize}
    \item We propose PolyNet, a novel neural network method with a continuous convolution operation invariant to the number of adjacent vertices, their permutations, and their pairwise distances in 3D shapes.
    
    \item We employ PolyNet on PolyShape, a polygon mesh representation with a multi-resolution structure that enables us to use a pooling operation named PolyPool.
    
    \item We achieve an improvement in classification and retrieval tasks compared to the previous mesh-based methods on the ModelNet dataset and the best classification performance on the 75 Superpixel MNIST.
 \end{itemize}

\section{Related work}
In this section, we review the related works based on the representation of the input 3D shapes: point cloud, voxel grid, 2D projection, and polygon mesh.
\\
\textbf{Point cloud.}
PointNet~\cite{Qi2016PointNetDL}, as a simple and effective DNN-based method on point clouds,  learns the features directly from each point and aggregate them as one global representation.  However, extracting local structures is important for the success of convolutional architectures.
To overcome the lack of local structure of this method, PointNet++~\cite{Qi2017PointNetDH} proposes a hierarchical neural network that employs PointNet~\cite{Qi2016PointNetDL} on the group of points divided into overlapping local patches.
Te \etal~\cite{Te:2018:RRG:3240508.3240621} and Wang \etal~\cite{10.1007/978-3-030-01225-0_4} utilized the GraphCNNs to learn the features from a local graph formed by the connection of the adjacent points. These graph-based methods produce little shape information since they do not explicitly represent the local neighboring points in an ordered alignment. 
%%%%%%%%%%%%%%%%%%%%%%%%%%%%%%%%%%%%%%%%%%%%%%%%%%%%%%%%
\\
\textbf{Voxel grid.} 3D ShapeNets~\cite{7298801} and VoxNet~\cite{7353481} transfer a 3D shape to a structured binary 3D grid called voxel gird. Then they learn the global features from the voxels by extending the CNN architectures from 2D to 3D convolutions. To reduce the computational complexity on the sparse voxels, Riegler \etal~\cite{Riegler2016OctNetLD} and Wang \etal~\cite{Wang:2017:OOC:3072959.3073608} applied the octree data structure.
However, these methods require heavy computations and unnecessarily voluminous data.
%%%%%%%%%%%%%%%%%%%%%%%%%%%%%%%%%%%%%%%%%%%%%%%%%%%%%%%%
\\
\textbf{2D projection.}
MVCNN~\cite{su15mvcnn} and RotationNet~\cite{8578624}
learn the features of a 3D shape over a multi-view rendered 2D images based on conventional 2D CNNs.
Moreover, pooling operations aggregate these feature values to reduce the rotation effects of the 3D shape \cite{7273863,Furuya:2017:DSH:3095140.3095148,su15mvcnn}.
However, these pooling operations lose a lot of geometric details among the views, such as two surfaces occluded to each other. 
SeqViews2SeqLabels~\cite{8453813} and SPNet\_VE~\cite{Yavartanoo2018SPNetD3} aggregate the information among the sequential views by considering the view specific importance to prevent the lost information.
%%%%%%%%%%%%%%%%%%%%%%%%%%%%%%%%%%%%%%%%%%%%%%%%%%%%%%%%
On the other hand, DeepPano~\cite{7273863} and PANORAMA-NN~\cite{3dor.20171045} consider a panoramic view of the 3D shape. They project the shape into a cylinder surrounding it to accumulate the contents of multiple views altogether.
To extend the number of viewpoints, utilizing a sphere instead of the cylinder leads CNNs to cover all views and learn more robust features consistent with rotations \cite{DBLP:journals/corr/abs-1801-10130, Yavartanoo2018SPNetD3}.
However, these image-based methods suffer from self-occlusions.
%%%%%%%%%%%%%%%%%%%%%%%%%%%%%%%%%%%%%%%%%%%%%%%%%%%%%%%%
\\
\textbf{Polygon mesh.} A polygon mesh is a discrete representation of the surface of a 3D shape with faces and vertices. This representation can be expressed as a graph; accordingly, any graph-based methods can be applied to it. The existing graph-based methods are classified into two main categories: spectral methods~\cite{Bruna2013SpectralNA,DBLP:journals/corr/HenaffBL15,NIPS2016_6081,Kipf2016SemiSupervisedCW,Levie2019CayleyNetsGC} and spatial methods~\cite{4773279,Atwood2016DiffusionConvolutionalNN,Niepert2016LearningCN,Gilmer2017NeuralMP, SplineConv,Masci:2015:GCN:2919341.2920992,8100059,texturenet}.
%%%%%%%%%%%%%%%%%%%%%%%%%%%%%%%%%%%%%%%%%%%%%%%%%%%%%%%%
The convolution operation in the spectral domain is defined by the eigendecomposition of the graph Laplacian, where the eigenvectors are the same as the Fourier basis~\cite{Bruna2013SpectralNA}. This process is basis-dependent, which indicates applying the learned parameters producing different features on a new domain~\cite{6579600}.
%\textcolor{Red}{
%Eynard \etal~\cite{7053905} proposed a compatible orthogonal basis, which is synchronized over various domains to overcome the basis dependency. However, finding the correspondence between two meshes is challenging due to requiring the correspondence information between the domains.
%}
Moreover, this operation is non-localized filtering in the spectral domain~\cite{NIPS2016_6081}. 
An efﬁcient method to solve the non-localization problem is approximating the local spectral filters via the Chebyshev polynomial expansion~\cite{NIPS2016_6081}.
%%%%%%%%%%%%%%%%%%%%%%%%%%%%%%%%%%%%%%%%%%%%%
On the other hand, there is no easy way to induce the weight sharing across different locations of the graph due to the difficulty of matching local neighborhoods in the spatial domain~\cite{Bruna2013SpectralNA}. 
Nevertheless, Atwood and Towsley~\cite{Atwood2016DiffusionConvolutionalNN} proposed a spatial filtering method that assumes information is transferred from a vertex to its adjacent vertex with a specific transition probability.
The power of the transition probability matrix implies that
farther adjacent vertices provide little information for the central
vertex.
Furthermore, Geodesic CNN~\cite{Masci:2015:GCN:2919341.2920992}, MoNet~\cite{8100059}, and SplineCNN~\cite{SplineConv} deal with the weight sharing problem by designing local coordinate systems for the central vertex in a local patch.
They apply a set of weighting functions to aggregate features on adjacent vertices. Then they compute a learnable weighted average of these aggregated features as the spatial convolution.
However, these methods are computationally expensive and require predefined local systems of coordinates. Moreover, %MoNet~\cite{8100059} on semi-regular polygon mesh representation that almost all of the vertices have the same degree,  gives the same weights to the relative position between each two nodes, due to using fixed local polar pseudo-coordinates around each node.    
\begin{figure*}[t]
\begin{center}
%\framebox[4.0in]{$\;$}
%\fbox{\rule[-.5cm]{0cm}{4cm} \rule[-.5cm]{12cm}{0cm}}
\includegraphics[width=1\textwidth]{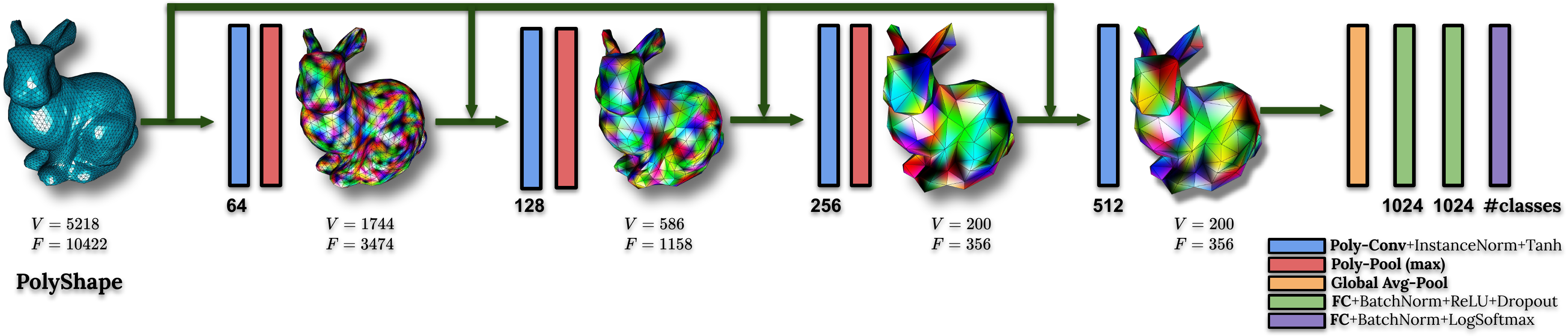}
\end{center}
\caption{The overview of PolyNet architecture. PolyNet takes a PolyShape as an input and applies four PolyConv followed by instance normalization and pooling layers and three fully connected (FC) followed by batch normalization layers to learn the local and the global features of the shape. 
%The first three layers of PolyConv are followed with PolyPool and the last is with a global average pooling layer. 
Then, we employ the hyperbolic tangent and ReLU activation functions to empower PolyConv and FC layers, respectively.
Moreover, PolyPool layers reduce the spatial dimensions and minimize the overfitting by utilizing the multi-resolution structure of the PolyShape. The global average pooling layer avoids the permutation ambiguities of the vertices. Note that $V$ and $F$ refer to the number of vertices and faces for each shape, respectively.}
\label{polynet}
\end{figure*} 
Neural3DMM~\cite{Bouritsas2019Neural3M} introduces the spiral convolution operation by enforcing a local ordering of vertices through the spiral operator.
An initial point for each spiral is a vertex with the shortest geodesic path to a fixed reference point on a template shape.
The remaining vertices of the spiral are ordered in the clockwise or counterclockwise directions inductively. 
However, finding a reference point for an arbitrary shape is challenging. Moreover, the initial point is not unique once two or more adjacent vertices have the same shortest path to the reference point.
%MeshNet~\cite{Feng2018MeshNetMN} solves the complexity and irregularity of traditional meshes based on face-unit and feature splitting.

\section{PolyNet}
In this section, we explain the details of our PolyNet architecture and its consistency to the number of adjacent vertices, their permutations, and their pairwise distances in 3D shapes.
PolyNet learns the features locally by \textbf{PolyConv} operation, and performs \textbf{PolyPool} procedure by utilizing the multi-resolution structure of our designed \textbf{PolyShape} representation. Figure~\ref{polynet} shows the overview of our PolyNet architecture.
A 3D shape with PolyShape representation passes through a straightforward network with three PolyConv layers followed by PolyPool layers and another PolyConv layer with a global average pooling to learn and extract the features. Then three fully connected layers classify the shape with these extracted features.
%%%%%%%%%%%%%%%%%%%%%sub_Section%%%%%%%%%%%%%%%%%%%%%%%%

\begin{figure*}[!t]
  \centering
\begin{tabular}{cc}
\includegraphics[page=1, width=8.3cm]{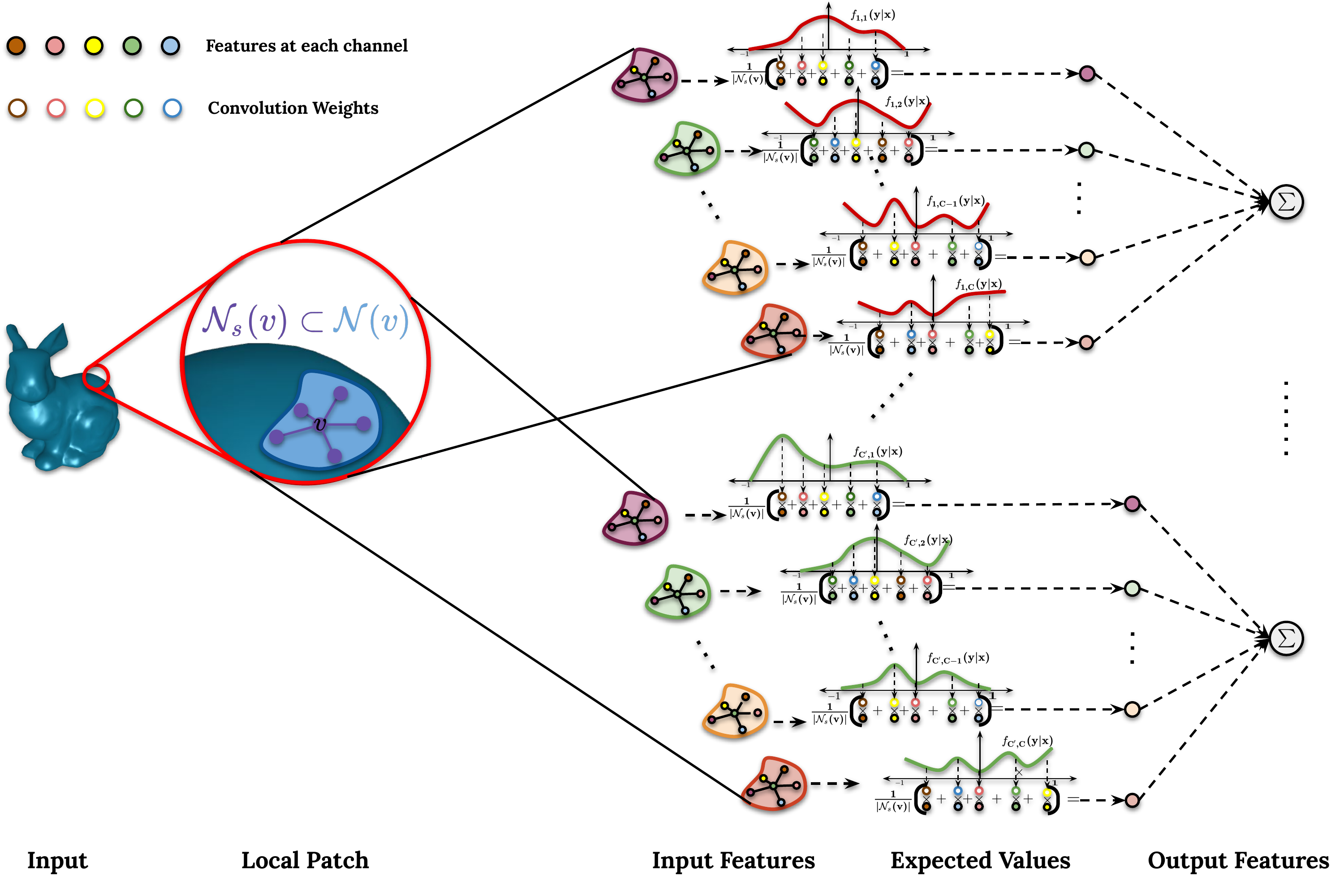}&
\includegraphics[page=2, width=8.3cm]{figures/PolyConv.pdf}\\
(a) Unsqueezed operations & (b) Squeezed operations\\\\
\end{tabular}
\caption{PolyConv operations over a local patch. A set of conditional PDFs approximated with polynomial functions with learnable coefficients is applied as the convolutional filters to learn the input features on a local patch of vertices $\mathcal{N}_s(v)\subset\mathcal{N}(v)$. (a) The unsqueezed operation includes $C\times C^{\prime}$ conditional PDFs as the convolutional filterers which map the input features to the higher dimensional output features. (b) The squeezed operation contains only $C$ conditional PDFs which is combined with a fully connected layer to map the input features to the higher dimensional output features.
Note that $C$ and $C^{\prime}$ refer to the size of input channels and output channels, respectively.}
\label{fig:Polyconv}
\end{figure*}

%%%%%%%%%%%%%%%%%%%%%%%Section%%%%%%%%%%%%%%%%%%%%%%%%%%

%%%%%%%%%%%%%%%%%%%%%sub_Section%%%%%%%%%%%%%%%%%%%%%%%%
\subsection{Polynomial convolution operation}\label{sect_Polyconv}
To overcome the challenges of the weight sharing across different vertices in the conventional CNNs and GraphCNNs, we propose PolyConv operation, which learns a probability density function (PDF) as a convolutional filter.
Let us assume that the surface of a 3D shape is a differential manifold $\mathcal{M}$. 
For a point $v$ and its neighbor $u$ in a local patch $\mathcal{N}(v)$ on the manifold $\mathcal{M}$, we define signals $x:\mathcal{M} \rightarrow [-1,1]$ and $y:\mathcal{M} \rightarrow [-1,1]$ as the features at those points, respectively.
Without loss of generality, we consider the convolutional weights of the standard CNN as the probability distributions. We then argue that a patch operation $D(v)$ in the standard CNN can be expressed as an expected value over the features on sample points $\mathcal{N}_s(v)\subset\mathcal{N}(v)$ surrounding $v$ as Eq.~\ref{implicit_1}:
\begin{equation}\label{implicit_1}
\begin{split}
    D(v)=\mathbb{E}[y|x]=\sum_{u\in\mathcal{N}_s(v)}w(u)y(u),
\end{split}
\end{equation}
where $w(u)$ is the corresponding probability (weight) to the point $u$. 
However, in a general graph or polygon mesh, the locations of the adjacent points are in a continuous domain and can vary; hence, it is not possible to assign a discrete distribution as the weights.
Therefore, we assume that there is an unknown conditional PDF that can express the convolution filter weights, and then we can formulate the expected value over each patch as Eq.~\ref{implicit_2}:
\begin{equation}\label{implicit_2}
\begin{split}
    D(v)=\mathbb{E}[y|x]=\int_{\mathcal{N}(v)}yf(y|x)dy,
\end{split}
\end{equation}
where $f(y|x)$ is the conditional probability of the feature $y$ on point $u$ in the neighborhood of the central point $v$ with the given feature $x$.
The conditional probability $f(y|x)$ can be written as Eq.~\ref{implicit_3}:
\begin{equation}\label{implicit_3}
\begin{split}
    f(y|x) &= \frac{f(x,y)}{f_x(x)}=\frac{f(x,y)}{\int_{-1}^{1}f(x,y)dy},
\end{split}
\end{equation}
where $f_{x}(x)$ is the marginal distribution that can be obtained by integrating the joint probability distribution, $f(x,y)$, over $y$.
Note that, since the value of feature $y$ is defined in $[-1, 1]\in\mathbb{R}$,
$\int_{-1}^{1}f(x,y)dy$ is a definite integral on the interval $[-1, 1]\in\mathbb{R}$.
We reformulate $f(y|x)$ by approximating $f(x,y)$ with a polynomial function of $x$ and $y$ by considering a certain degree $d$ as Eq.~\ref{implicit_4}:
\begin{equation}\label{implicit_4}
\begin{split}
    f(y|x) &=\frac{\sum\limits_{0\leq i,j,i+j\leq d} a_{i,j}x^i y^j}{\sum\limits_{0\leq i\leq d} b_{i}x^i},
\end{split}
\end{equation}
where the coefficients $b_i$ can be directly obtained by computing the marginal distribution $f_x(x)$ from $f(x,y)$.
To ensure that the polynomial function as a PDF is always positive, the coefficient matrix $A$ in the compact form of the polynomial function as Eq.~\ref{implicit_5} must be positive definite.
\begin{equation}\label{implicit_5}
\begin{split}
    f(x,y) &= \sum\limits_{0\leq i,j,i+j\leq d} a_{i,j}x^i y^j\\  &= X^{T}AX>0,
\end{split}
\end{equation}
where $X$ is the vector of variables $x$ and $y$ with degrees of less or equal $d/2$. 
Therefore, instead of learning the coefficient matrix $A$, we parameterize it as $A=B B^{T}\succ0$.
Indeed, we approximate the conditional PDF $f(x|y)$ as the continuous convolutional filters with the polynomial functions, which are parameterized by the learnable symmetric matrix $B$. For more details, refer to the supplementary material.
The large degree of freedom in polynomial functions allows approximating any complex distributions.

It is important to note that since we only have few samples $\mathcal{N}_{s}$ (e.g., points, vertices, etc.) in each local patch $\mathcal{N}$ on the manifold $\mathcal{M}$, computing the exact expected value over each local patch is not possible with Eq.~\ref{implicit_2}.
Therefore, we approximate the integral by taking the weighted average over these sample points as Eq.~\ref{implicit_6}:
\begin{equation}\label{implicit_6}
\begin{split}
\int_{\mathcal{N}(v)}yf(y|x)dy  \simeq
\frac{1}{|\mathcal{N}_{s}(v)|}\sum_{u\in \mathcal{N}_{s}(v)} yf(y|x).
\end{split}
\end{equation}

Finally, we design unsqueezed and squeezed operations based on the proposed patch operator as the convolutions to learn from the input features, as shown in Figure \ref{fig:Polyconv}.
In the first approach, similar to the conventional CNNs, we consider multiple conditional PDFs corresponding to the input and the output channels as the convolution filters.
However, this operation requires heavy computations and large memory usage.
Therefore, we squeeze it by allocating different conditional PDFs to only the input channels and aggregate the results by a fully connected layer.
The second approach is beneficial when the number of input vertices is large.
Therefore, with this continuous convolution operation, we can locally learn the features from the surface of 3D shapes, which is invariant to the number of vertices in a local patch, their permutation, and their pairwise distances.

\begin{figure*}[t]
\begin{center}
%\framebox[4.0in]{$\;$}
%\fbox{\rule[-.5cm]{0cm}{4cm} \rule[-.5cm]{12cm}{0cm}}
\includegraphics[width=1\textwidth]{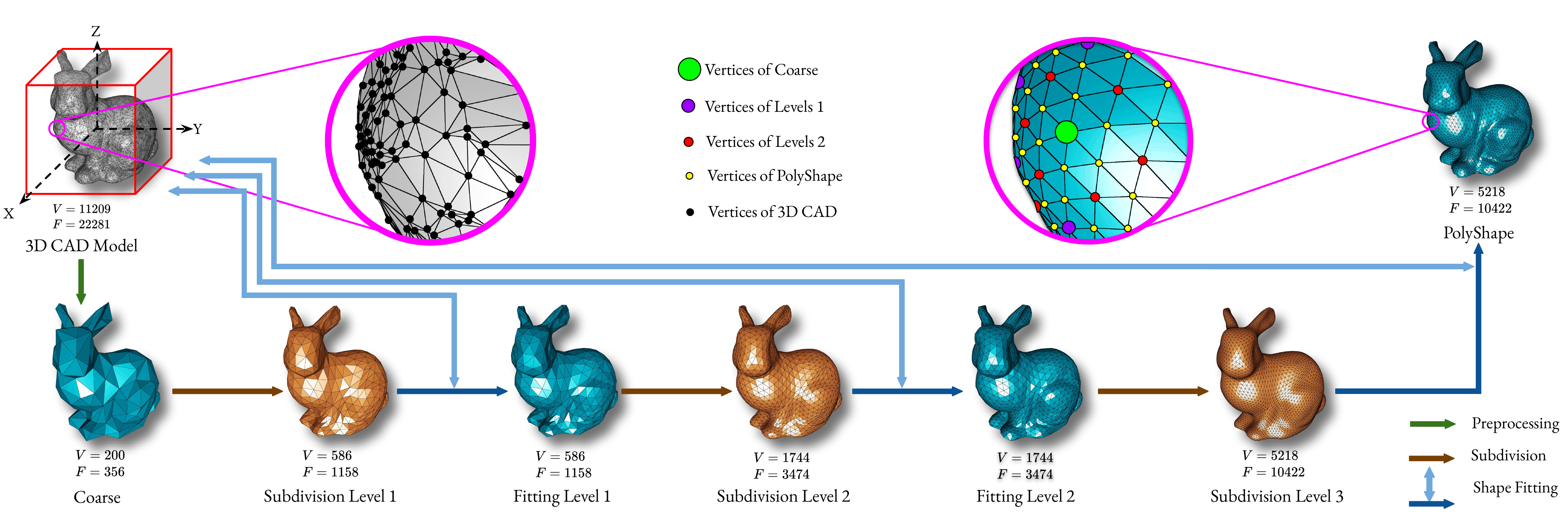}
\end{center}
\caption{The overview of PolyShape processing. A given 3D CAD model passes through a preprocessing pipeline to produce a coarse polygon mesh with a simpler topology and single connected component. 
Next, the subdivisions and shape fitting procedures sequentially create PolyShape with the multi-resolution structure for the shape.  %Note that $V$ and $F$ refer to the number of vertices and faces for each shape, respectively.
}
\label{polyshape}
\end{figure*}

\subsection{Polygonal shape representation and pooling}
To apply pooling after each convolution operation, we present PolyShape representation with a multi-resolution structure made of a sequence of the subdivisions and shape fittings, as shown in Figure~\ref{polyshape}.
This multi-resolution structure enables the pooling operations without any learnable parameter, which is similar to the multi-level pooling on images.
Moreover, PolyShape maintains the structural details and the topology of the shape after each pooling and provides a semi-regular structure that benefits the analysis of the local structure of the shape (i.e., each vertex and its corresponding neighborhood)~\cite{payan:2015:semi}.
For PolyShape processing, we first employ the mesh fusion~\cite{Stutz2018ARXIV} to a given 3D CAD model for the abstraction of the shapes with a simpler topology.
Next, we fix the geometric errors of the meshes, such as the non-manifold edges and double vertices, and then reduce the number of the vertices to obtain a coarse mesh with nearly $400$ vertices. 
Lastly, we subdivide the coarse mesh and fit the resulting mesh to the given model to restore the details of the original shape.
We apply this sub-division routine iteratively, as frequent as the number of the pooling layers in the PolyNet (i.e., $3$ times).
For the subdivision, there are two common methods: the primal triangle quadrisection (PTQ)~\cite{loop:1987:smooth} and the $\sqrt{3}$-subdivision~\cite{Kobbelt:2000::344779.344835}.  
PTQ is a straightforward approach that splits a triangle into four sub-triangles. 
It creates new vertices on each edge in the original mesh and connects them to each of the other new vertices from the same face.
The other strategy, $\sqrt{3}$-subdivision, adds the new vertices inside each triangle in the original mesh and connects the new vertices to each of its three old surrounding vertices and adjacent new vertices. 
Every two iterations of the $\sqrt{3}$-subdivision separate each original triangle into nine sub-triangles. 
Thus, PolyShapes have fewer triangles by the $\sqrt{3}$-subdivision than the PTQ. 
We evaluate the effectiveness of the PolyShapes made by both subdivisions in Section~\ref{sec:experiment}.

\begin{figure}[!t]
  \centering
\begin{tabular}{ccc}
\includegraphics[width=8.2cm]{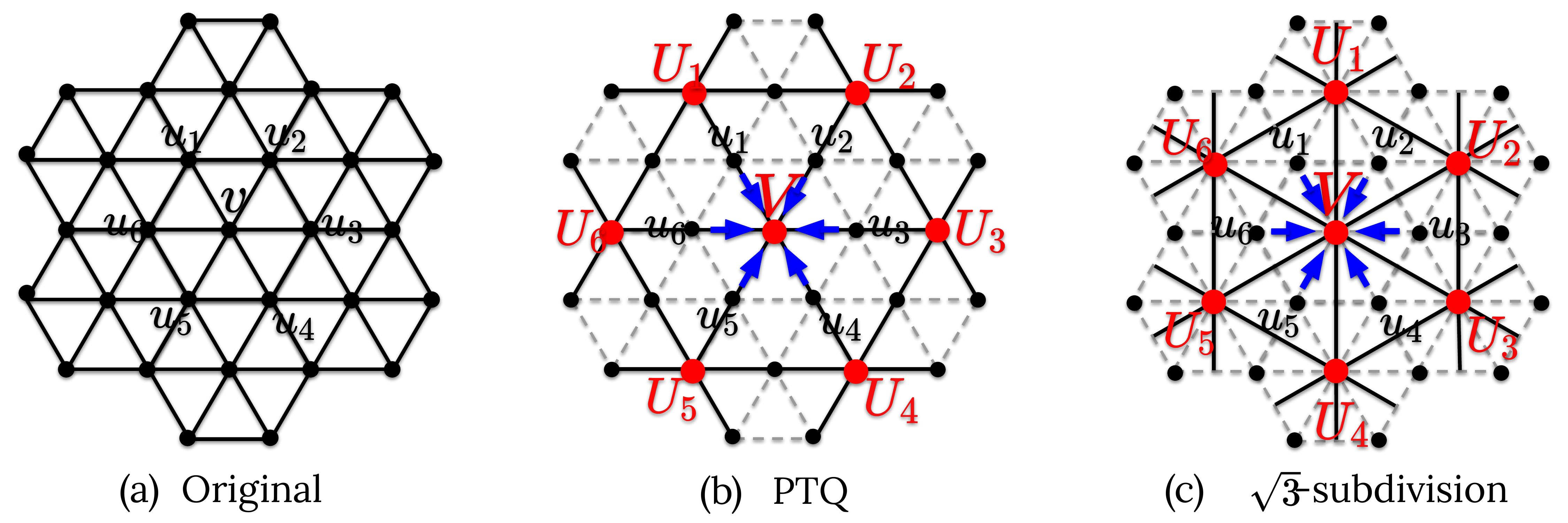}
\end{tabular}
\caption{PolyPool operations. The black dots and red dots show the vertices before and after PolyPool, respectively. Dash lines indicate the omitted edges after the subdivisions.}
\label{fig:pool}
\end{figure}

With the multi-resolution structure of PolyShape, we can downsample the output of PolyConv layers by collapsing the neighboring vertices to each interior vertex (i.e., the vertices of the coarser meshes), shown in Figure~\ref{fig:pool}. 
The PTQ and $\sqrt{3}$-subdivision upsample the mesh vertices by adding another vertex at the center of each edge and each triangle, respectively. The downsampling procedures are accomplished as the inverse process of the upsampling methods, which allow us to generate relatively larger polygons, as shown in Figure~\ref{fig:pool}. Therefore, we can reduce the number of polygons by the factors of four and three by employing the PTQ and $\sqrt{3}$-subdivision methods, respectively. We use these downsampling procedures as the pooling, which facilitates aggregating the features, where each vertex $V$ on the downsampled mesh takes the maximum (max-pool) over features values of the vertices $v\cup\{u_i\}_{i=1}^{m}$ in a local patch on the output of PolyConv layers.
Therefore, PolyPool enables describing a 3D shape with a large number of vertices by aggregating the features. These aggregated features are much lower in dimension compared to using all of the extracted features and also can improve the performance like the conventional pooling operation of the 2D CNNs~\cite{6144164}.

%
%\begin{figure}
%\begin{center}
%\includegraphics[width=.85\textwidth]{figures/remeshing_framework.pdf}
%\end{center}
%\caption{The overview of the remeshing process based on the Houdini.}
%\label{fig:remeshing_framework}
%\end{figure}

%%%%%%%%%%%%%%%%%%%%%sub_Section%%%%%%%%%%%%%%%%%%%%%%%%

%%%%%%%%%%%%%%%%%%%%%%%%%%%%%%%%%%%%%%%%%%%%%%%%%%%%%%%%

%%%%%%%%%%%%%%%%%%%%%%%%%%%%%%%%%%%%%%%%%%%%%%%%%%%%%%%%

%%%%%%%%%%%%%%%%%%%sub_sub_Section%%%%%%%%%%%%%%%%%%%%%%

%%%%%%%%%%%%%%%%%%%sub_sub_Section%%%%%%%%%%%%%%%%%%%%%%

%%%%%%%%%%%%%%%%%%%%%%%%%%%%%%%%%%%%%%%%%%%%%%%%%%%%%%%%%%%%%%%%%%%%%%%%%%%%%%%%%%%%%%%%

%%%%%%%%%%%%%%%%%%%%%%%%%%%%%%%%%%%%%%%%%%%%%%%%%%%%%%%%%%%%%%%%%%%%%%%%%%%%%%%%%%%%%%%%%%%%

%\subsubsection{Shape Descriptors}\label{sect_Polypool}
%Furthermore, we designed an unsupervised key-point detection to aggregate the most relevant features on %the coarse surface for both classification and retrieval task, as shown in %Figure.~\ref{normal_features}.
%After the last convolution layer, we compute the L1 norm of features of the vertices on the coarse %surface and select the $N$ vertices with the largest norms as the key-points and their features as the %shape descriptors.
%Finally, we pass the average features over all vertices of a 3D shape concatenated with its shape %descriptors to multiple fully connected layers for classifying the shape.

\begin{figure*}[!t]
  \centering
\begin{tabular}{cc}
\includegraphics[page=1, width=8cm]{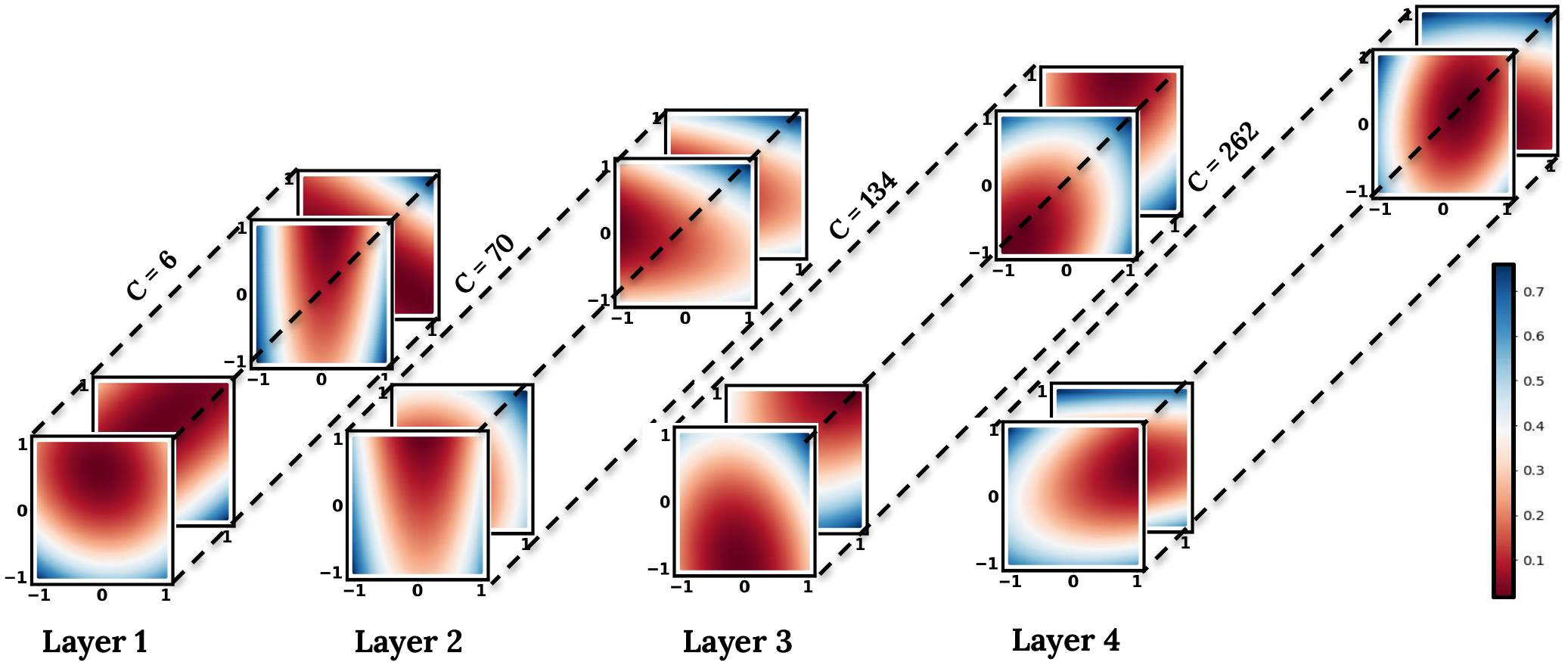}&
\includegraphics[page=2, width=8cm]{figures/shapes/Distributions_conv.pdf}  \\
(a) Joint distributions & (b) Marginal distributions\\
\end{tabular}
\caption{Joint and marginal distributions. Visualization of (a) the learned joint distributions $f(x,y)$ and (b) the marginal distributions $f_x(x)$ approximated by polynomial functions of degree $d=2$ on the ModelNet-10 with the $\sqrt{3}$-subdivision.}
\label{fig:distribution}
\end{figure*}

\section{Experiments}\label{sec:experiment}

%%%%%%%%%%%%%%%%%%%%%%%%%%%%%%%%%%%%%%%%%%%%%%%%%%%%%%%%%%%%%%%%%%%%%%%%%%%%%%%%%%%

%%%%%%%%%%%%%%%%%%%%%%%%%55%%%%%%%%%%%%%%%%%%%%%%%%%%%%
In this section, we present the details of the datasets and several experiments of PolyNet on the 3D shapes with and without PolyShape representation and graph representation of images.
We compare  our proposed method with the state-of-the-art methods on both classification and retrieval tasks.
We use Adam optimizer in all of our experiments with the initial learning rate as 1.e-3 and 1.e-2 for unsqueezed and squeezed cases, respectively. We set a mini-batch size as 100 and 10 for experiments on 3D shapes and graph representation of images, respectively.
We choose hyperbolic tangent for the activation functions on PolyConv layers to guarantee that the input features to the next layers are in the interval $[-1,1]\in\mathbb{R}$, and we use cross-entropy loss between the model predicted scores and ground truth labels. We also implement our model in Python3.6 using PyTorch via CUDA instruction. 
PolyShape processing, including both subdivision methods, takes 92 ms for one CAD model on average, and we ensure all the conversions are successful for the ModelNet dataset.
The average testing times for the baseline of PolyNet per shape for the $\sqrt{3}$-subdivision and PTQ are 13 ms and 18 ms, respectively. 
We will publish the code for both PolyShape processing and PolyNet.

%%%%%%%%%%%%%%%%%%%%%%%%%%%%%%%%%%%%%%%%%%%%%%%%%%%%%%%%%%%%%%%%%%%%%%%%%%%%%%%%

%%%%%%%%%%%%%%%%%%%%%%%%%%%%%%%%%%%%%%%%%%%%%%%%%%%%%%%%%%%%%%%%%%%%%%%%%%%%%%%%%%%%%%%%
\subsection{Datasets}\label{data}
In our experiments, we use both the ModelNet-10 and the ModelNet-40 datasets~\cite{7298801} containing 4,899 CAD models (3991 for training and 908 for testing) in 10 categories and 12,311 CAD models (9843 for training and 2468 for testing) in 40 categories, respectively.
We apply our PolyShape processing to the CAD models with Houdini~\cite{houdini}, a popular 3D modeling software.
For more details about the PolyShape pipeline, refer to supplemental material.
Additionally, we translate and scale the resulting PolyShapes into the bounding box $[-1,1]^3\in \mathbb{R}^3$.
We extract the coordinates $(x,y,z)\in \mathbb{R}^3$ and the normal vectors $(nx,ny,nz)\in \mathbb{R}^3$,  for all vertices as the first input into PolyNet. 
%%%%%%%%%%%%%%%%%%%%%%%%%%%%%%%%%%%%%%%%%%%%%%%%%%%%%%%%%%%%%%%%%%%%%%%%%%%%%%%%%%%%%%%%%%%%
We also use the MNIST dataset~\cite{726791}, which includes $28\times28$ images. These images are represented as different graphs so that each vertex and each edge corresponds to a superpixel and the spatial relation between two superpixels, respectively~\cite{8100059}. Therefore, we consider the construction of superpixel-based graphs with 75 vertices.
We use the standard splitting of the MNIST dataset, including 60k and 10K images for training and testing, respectively.

\begin{table}[t]
\small
\begin{center}
\setlength\tabcolsep{0.7pt} % default value: 6pt
\begin{tabular}{|l|c|c|c|c|c|c|c|c|c|}
\hline
\multirow{2}{*}{\bf Conv.}
&\multicolumn{3}{c|}{\bf $\sqrt{3}$-subdivision }
&\multicolumn{3}{c|}{\bf PTQ} & \multirow{2}{*}{\bf Params.}\\
\cline{2-7}
 & max & avg & Time & max & avg & Time & \\
%\cline{3-6}
\hline\hline
XConv~\cite{XConv}&{84.58}&{83.31}& 173ms & {85.54}&{83.85} & 835ms & 473k\\
SplineConv~\cite{SplineConv}&{93.46}&{92.72} & 12ms &{93.16}&{92.66} & 18ms & 111m \\
ChebConv~\cite{ChebConv}&{93.95}&{93.42} & 12ms &{93.70}&{93.11} & 14ms & 712k\\
GCNConv~\cite{GCNConv}&{93.85}&{93.38} & \textbf{9ms} &{93.85}&{93.30} & \textbf{13ms} & \textbf{179k} \\
GMMConv~\cite{8100059}&{93.32}&{92.68} & 17ms &{93.20}&{92.36} & 28ms & 4.6m \\
FiLMConv~\cite{FiLMConv}&{94.43}&{93.89} & \textbf{9ms} &{94.30}&{93.76} & \textbf{13ms} & 1.1m \\
\hline
PolyConv($d=4$)&{93.96}&{93.13} & 25ms &{94.00}&{92.38}  & 38ms & 189k \\
PolyConv($d=2$)&\textbf{94.52}&{\textbf{93.95}} & 13ms &{\textbf{94.40}}&{\textbf{94.08}}  & 18ms & 182k\\
\hline
\end{tabular}
\end{center}
\caption{Classification accuracy (Acc\%), testing time, and number of parameters in only convolution layers on the ModelNet-10 with both subdivision strategies for the various convolution operations in PolyNet.}
\label{table:convs}
\end{table}

%%%%%%%%%%%%%%%%%%%%%%%%%%%%%%%%%%%%%%%%%%%%%%%%%%%%%%%%%%%%%%%

%%%%%%%%%%%%%%%%%%%%%%%%%%%%%%%%%%%%%%%%%%%%%%%%%%%%%%%%%%%%%%%%%%%%%%%%%%%%%%%%%%
\subsection{Convolution operation}
We evaluate our convolution operation PolyConv with different configurations and compare it with various famous convolutional operations as shown in Table~\ref{table:convs}.
Since squeezed PolyConv requires fewer computations and less memory usage compared to the unsqueezed version due to the less number of learnable parameters, we use it for the experiments of 3D shape classification where the inputs are extremely large (roughly 10k vertices).
We consider two different degrees $d=2$ and $d=4$ for each polynomial function defined in Eq.~\ref{implicit_5} which each requires six and 21 learnable coefficients for a patch operation, respectively.
The results on MoldelNet-10 with both subdivision strategies show that PolyConv with degree $d=2$ achieves relatively higher performance than degree $d=4$, which can be due to its straightforward and easier to learn structure for approximating the distributions.
%We evaluate and compare the classification accuracy results on the polynomial functions with different degrees on the ModelNet-10 and demonstrate that increasing the degree $d$ in polynomial functions can improve the accuracy for both pooling strategies based on the PTQ and $\sqrt{3}$-subdivision.
Furthermore, we evaluate PolyNet on ModelNet-10 by replacing PolyConv with well-known convolutions such as ChebConv~\cite{ChebConv}, GCNConv~\cite{GCNConv}, GMMConv~\cite{8100059}, SplineConv~\cite{SplineConv}, XConv~\cite{XConv}, and FiLMConv~\cite{FiLMConv}.
We show that PolyConv with degree $d=2$ achieves superior performances compared to all mentioned convolutions for both subdivision strategies.
Towards a better understanding of the PDFs, we visualize the learned joint PDFs $f(x,y)$ and the marginal PDFs $f_x(x)$ of squeezed PolyConv for polynomial functions of degree $d=2$ which are learned on the ModelNet-10 dataset with the $\sqrt{3}$-subdivision in Figure~\ref{fig:distribution}.
The results illustrate the diversity of learned PDFs among different input channels and different layers of PolyNet.

%%%%%%%%%%%%%%%%%%%%%%%%%%%%%%%%%%%%%%%%%%%%%%%%%%%%%%%%%%%%%%%
\begin{figure}[!t]
  \centering
\includegraphics[width=8.5cm]{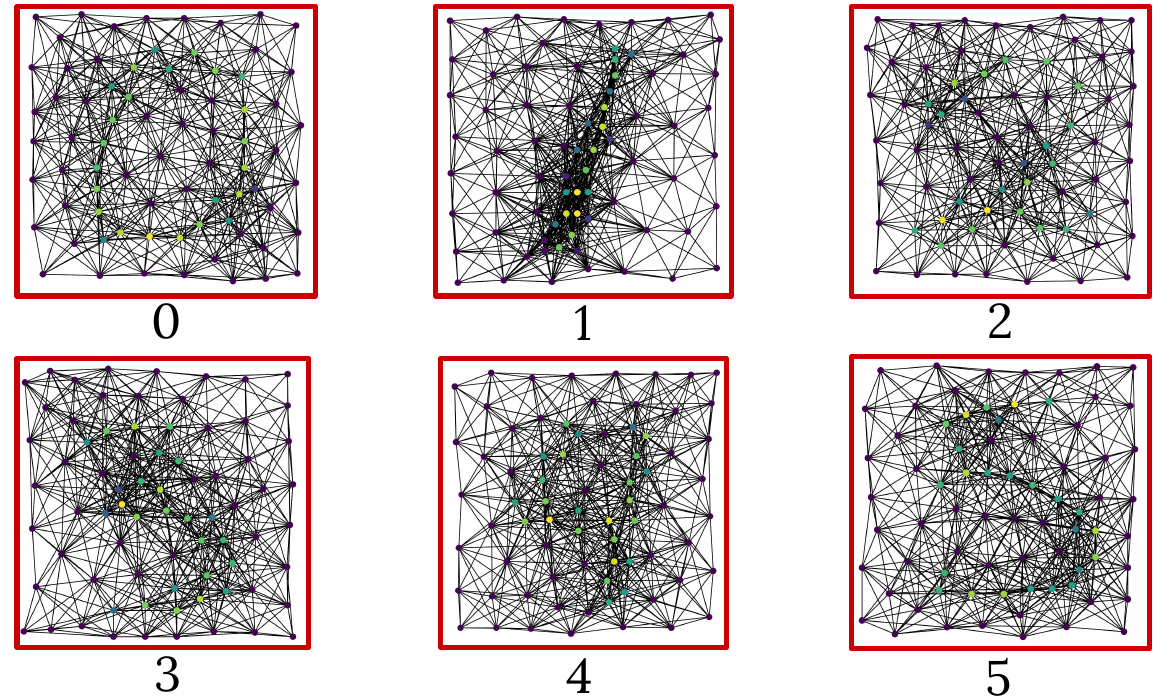}
\caption{The graphs of the MNIST dataset. Visualization of handwritten digits in the MNIST dataset individually represented as the graph of superpixel with 75 vertices.}
\label{fig:mnist}
\end{figure}

On the other hand, we evaluate our unsqueezed PolyConv and compare it with the squeezed PolyConv for the polynomial functions of degree $d=2$ on a classical task of handwritten digit classification in the graph representation of the MNIST dataset~\cite{8100059} with 75 vertices.
Despite the simplicity of underlying images, this is a challenging task due to the lack of a regular grid structure among the nodes, as shown in Figure~\ref{fig:mnist}.
We use three convolutional layers (256,256,256) and three fully connected layers (1024,1024,10) in the network architecture.
We employ both unsqueezed and squeezed PolyConvs as the convolution operations and the graclus clustering for the pooling procedure. We use the position information of the vertices as the extra features.
We show in Table~\ref{table:mnist} that PolyConv outperforms the existing methods, which demonstrates the strengths of PolyConv to learn features from irregular data as well as semi-regular data.
Moreover, we show that while the squeezed version of PolyConv achieves only slightly lower performance, it requires 135k learnable parameters in PolyConv layers, which is much more efficient than unsqueezed PolyConv with 792k parameters. Note that the average testing time for squeezed PolyConv on the samples of the 75 Superpixel MNIST is 1.4 ms, while it is 12.4 ms for the unsquzeed PolyConv.

\begin{table}[b]
\begin{center}
\begin{tabular}{|l|c|c|}
\hline
\bf Method  & \bf Acc. (max)  \\ 
\hline\hline 
MoNet~\cite{8100059} & 91.11 \\
SplineCNN~\cite{SplineConv} &  95.22   \\
GCGP~\cite{GCGP} &  95.80   \\
GAT~\cite{GAT} &  96.19   \\
PNCNN~\cite{PNCNN} & 98.76  \\
\hline
PolyConv  (squeezed) &  98.39 \\
PolyConv (unsqueezed)  &  \textbf{98.95 } \\
\hline
\end{tabular}
\end{center}
\caption{Classification accuracy (Acc\%) of various methods and our PolyConv operations on a superpixel representation of the MNIST dataset with 75 vertices.}
\label{table:mnist}
\end{table}

\begin{table}
\begin{center}
\setlength\tabcolsep{1.2pt} % default value: 6pt
\begin{tabular}{|l|c|c|c|c|c|c|}
\hline
\multirow{2}{*}{\bf Pooling}  & \bf Poly & \multicolumn{2}{c|}{\bf Accuracy} & \multirow{2}{*}{\bf Time} & \bf \multirow{2}{*}{Num.} \\ 
\cline{3-4}
& \bf Shape & max & avg & &\\
\hline 
\hline
No pooling & \xmark & {94.11} & {92.69} & 22ms &2.8k\\
Graclus~\cite{Graclus} & \xmark  & {94.14} & {92.73} & 16ms &2.8k \\
\hline
PolyPool(PTQ) & \cmark & {94.40}&\textbf{94.08} & 18ms &{25.7k}\\
PolyPool($\sqrt{3}$-sub) & \cmark & \textbf{94.52}  & {93.95}& 13ms  &10.8k\\
\hline
\end{tabular}
\end{center}
\caption{Classification accuracy (Acc\%), average testing time, and average number of vertices for various types of poolings and data with and without PolyShape processing.}
\label{table:pooling}
\end{table}

%%%%%%%%%%%%%%%%%%%%%%%%%%%%%%%%%%%%%%%%%%%%%%%%%%%%%%%%%%%%%%%%%%%%%%%%%%%%%%%%%%%%%%%%
\subsection{Pooling layers}
To show the benefits of our PolyPool operation, we apply PolyNet with various configurations of the pooling operation and input data type and compare the results in Table~\ref{table:pooling}.
We consider two different data representations, including data with and without PolyShape processing for the ModelNet-10 dataset.
Our experiments demonstrate that PolyPool with PolyShape representation can effectively improve the performance, especially by the pooling based on $\sqrt{3}$-subdivision.
Moreover, we employ a three-level graclus~\cite{Graclus} as an efficient clustering algorithm on the data without PolyShape representation. However, the results show lower accuracy when we use pooling based on the graclus clustering compared to both $\sqrt{3}$-subdivision and PTQ. We interpret the accuracy gap as the effect of losing structural information of 3D shapes by applying the graclus clustering.
Note that we compute the maximum value (max-pool) of each local patch for all pooling strategies.

%%%%%%%%%%%%%%%%%%%%%%%%%%%%%%%%%%%%%%%%%%%%%%%%%%%%%%%%%%%%%%%%%%%%%%%%%%%%%%%%%%

\subsection{3D shape classification}
To improve the 3D shape classification performance, we combine the output of the last PolyConv layer for both subdivisions by taking an average over their features. 
We compare the classification results of our PolyNet, with the recent state-of-the-art methods in Table~\ref{table:sota} on the ModelNet-10 and ModelNet-40 datasets.
We note that our PolyNet outperforms all the mesh-based and voxel-based approaches on the classification task and achieves comparable performance to the methods based on point clouds. 
The performance gaps between the 2D projection-based methods and other methods are due to utilizing pre-trained networks on a large number of images. Moreover, images include texture information produced by lights and shadows, while the other representations suffer from a lack of such information.                                                                                                                                                                                                                                                                                                                                                      

\begin{table}[t]
\small
\begin{center}
\setlength\tabcolsep{0.5pt} % default value: 6pt
\begin{tabular}{|c|l|c|c|c|c|c|}
\hline
\multirow{2}{*}{\bf Rep.} &\multirow{2}{*}{\bf Method}
&\multicolumn{2}{c|}{\bf ModelNet-10}
&\multicolumn{2}{c|}{\bf ModelNet-40}\\
\cline{3-6}
 & & Acc & mAP & Acc & mAP
\\ \hline\hline
\multirow{5}{*}{2D}       
& DeepPano~\cite{7273863} & {85.45} & {84.18} & {77.63} & {76.81}\\
\multirow{5}{*}{Projection}     
& MVCNN~\cite{su15mvcnn}& {-} & {-} & {90.10} & {79.50}\\
& PANORAMA-ENN~\cite{3dor.20171045} & {96.85} & {93.28} & {95.56} & {86.34}\\
%& MHBN~\cite{8578125} & - & & {95.00} & {-} & {94.70} & {-}\\
& SPNet\_VE~\cite{Yavartanoo2018SPNetD3} & {97.25} & {\bf 94.20} & {92.63} & {85.21}\\
%& SeqViews2SeqLabels~\cite{8453813}  & {94.82} & {91.43} & {93.40} & {\bf 89.09}\\
%& SeqViews2SeqLabels~\cite{Han2019SeqViews2SeqLabelsL3} &  & & {94.82} & {91.43} & {93.40} & {89.09}\\
%& Ma et al.~\cite{8490588} & - & & {95.29} & {93.19} & {91.05} & {84.34}\\
%& iMHL~\cite{8424480} & - &  & {-} & {-} & {97.16} & {-}\\
& RotationNet~\cite{8578624} & {\textbf{98.46}} & - &{\textbf{97.37}} & {-}\\
\hline
\multirow{5}{*}{voxel}    
& 3D ShapeNets~\cite{7298801} & {83.54} & {\bf 68.26} & {77.32} & {\bf 49.23}\\
\multirow{5}{*}{grid}    
& VoxNet~\cite{7353481}& {92.00} & {-} & {83.00} & {-}\\
& VRN~\cite{DBLP:journals/corr/BrockLRW16} & {93.61} & {-} & {91.33} & {-}\\
& FusionNet~\cite{Hegde2016FusionNet3O} & {93.11} & {-} & {90.80} & {-}\\
%& 3D-GAN~\cite{10.5555/3157096.3157106} & 11M & {91.00} & {-} & {83.30} & {-}\\
%& OctNet~\cite{Riegler2016OctNetLD} & - & {90.42} & {-} & {-} & {-}\\
& LP-3DCNN~\cite{DBLP:conf/cvpr/KumawatR19}  & {\textbf{94.40}} & {-} & {\textbf{92.10}} & {-}\\
\hline
\multirow{10}{*}{Point}  
& PointNet~\cite{Qi2016PointNetDL} & {-} & {-} & {89.20} & {-}\\
\multirow{10}{*}{cloud}  
& PointNet++~\cite{Qi2017PointNetDH} & {-} & {-} & {91.90} & {-}\\
& SO-Net~\cite{DBLP:journals/corr/abs-1803-04249} & \textbf{95.50} & {-} & {90.80} & {-}\\
& KCNet~\cite{DBLP:conf/cvpr/ShenFYT18}  & {94.40} & {-} & {91.00} & {-}\\
& PCNN~\cite{PCNN}  & {94.90} & {-} & {92.30} & {-}\\
& SpiderCNN~\cite{SpiderCNN}  & {-} & {-} & {92.40} & {-}\\
& PointCNN~\cite{NEURIPS2018_f5f8590c} & {-} & {-} & 92.50 & {-}\\
& DGCNN~\cite{PHAN2018533} & {-} & {-} & 92.90 & {-}\\
& KPConv~\cite{DBLP:journals/corr/abs-1904-08889} & {-} & {-} & 92.90 & {-}\\
& RS-CNN~\cite{inproceedings} &{-} & {-} & 
\textbf{93.60} & {-}\\
 
%& Point2Sequence~\cite{Liu2019Point2SequenceLT}  & {\bf 95.30} & {-} & {\bf 92.60} & {-}\\
\hline 
\multirow{9}{*}{Polygon} 
& SPH~\cite{Kazhdan2003RotationIS} & {79.79} & {44.05} & {68.23} & {33.26}\\
\multirow{9}{*}{Mesh} 
& Geometry Image~\cite{Sinha2016DeepL3} & {88.40} & {74.90} & {83.90} & {51.30}\\
& MeshNet~\cite{Feng2018MeshNetMN} & {-} & {-} & {91.90} & {81.90}\\
& Cross-atlas~\cite{8954385} & {91.20} & {-} & {87.50} & {-}\\
& SNGC~\cite{SNGC} & {-} & {-} & {91.60} & {-}\\
& MeshWalker~\cite{MeshWalker} & {-} & {-} & {92.30} & {-}\\
\cline{2-6}
& PolyNet ($\sqrt{3}$),(d=2) & 94.52 & 83.91 & 92.14 & 82.36\\
& PolyNet (PTQ),(d=2) & 94.40 & 83.84 & 92.06 & 81.91\\
& PolyNet (PTQ,$\sqrt{3}$),(d=2) & \textbf{94.93} & \textbf{84.62} & \textbf{92.42} & \textbf{82.86} \\
\hline
\end{tabular}
\end{center}
\caption{Classification accuracy (Acc\%) and mean Average Precision (mAP\%) of PolyNet compared to the state-of-the-art methods based on different representations on the ModelNet-10 and the ModelNet-40 datasets.}
\label{table:sota}
\end{table}

\begin{figure}
\centering
    \begin{subfigure}{0.19\linewidth}
        \includegraphics[width=\linewidth]{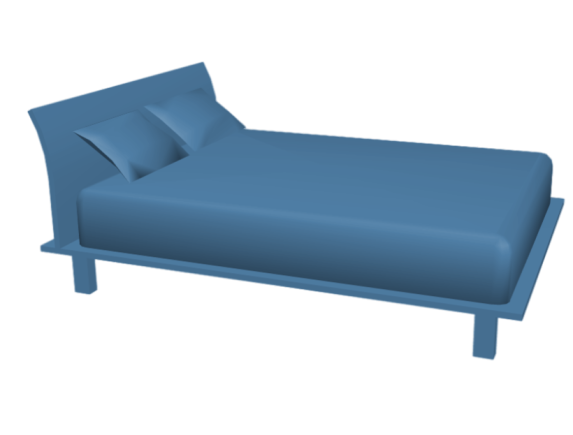}
    \caption*{}
    \end{subfigure}
\hfil
    \begin{subfigure}{0.19\linewidth}
        \includegraphics[width=\linewidth]{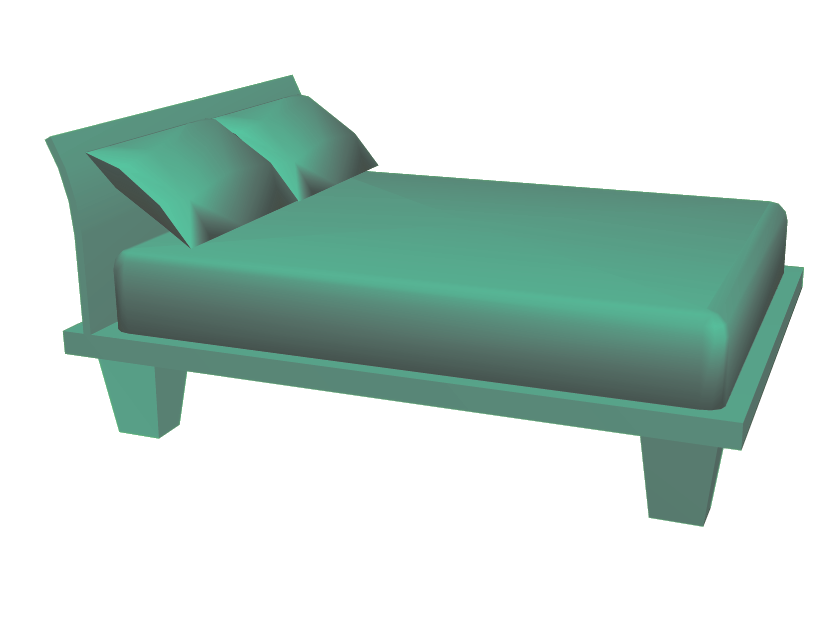}
    \caption*{}
    \end{subfigure}
\hfil
    \begin{subfigure}{0.19\linewidth}
        \includegraphics[width=\linewidth]{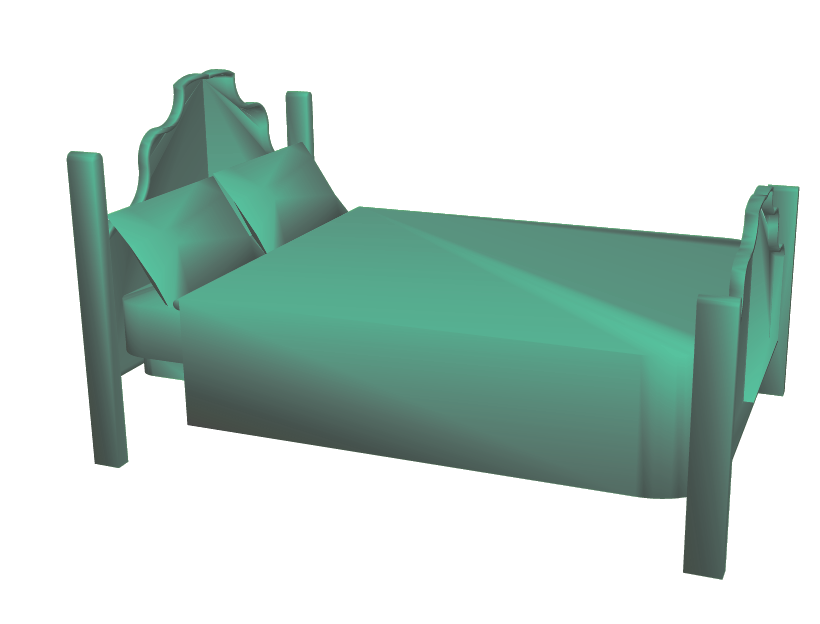}
    \caption*{}
    \end{subfigure}
\hfil
    \begin{subfigure}{0.19\linewidth}
        \includegraphics[width=\linewidth]{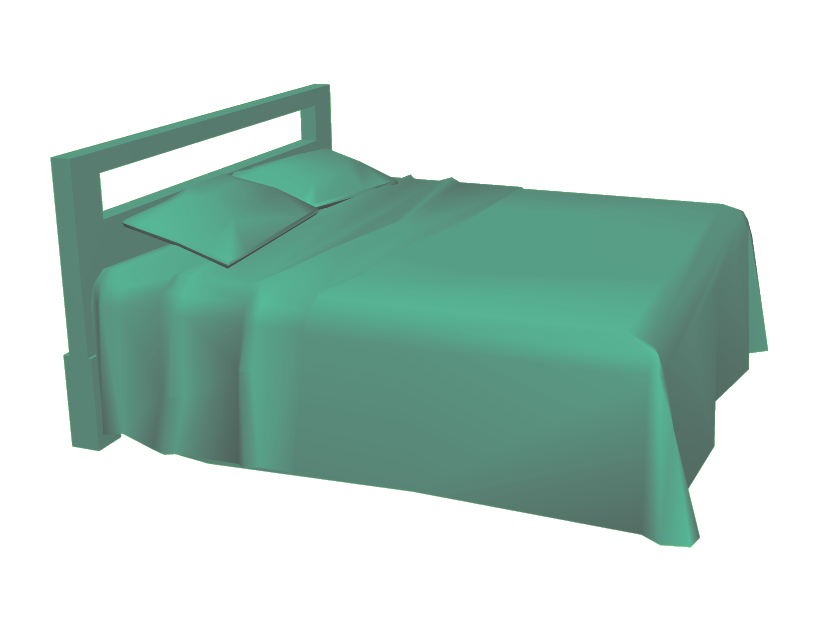}
    \caption*{}
    \end{subfigure}\\[-2em]
%%%%%%%%%%%%%%%%%%%%%%%%%%%%%%%%%%%%%%%%%%Bed%%%%%%%%%%%%%%%%%%%%%%%%%%%%%%%%%%%%%%%%%%%%
    \begin{subfigure}{0.19\linewidth}
        \includegraphics[width=\linewidth]{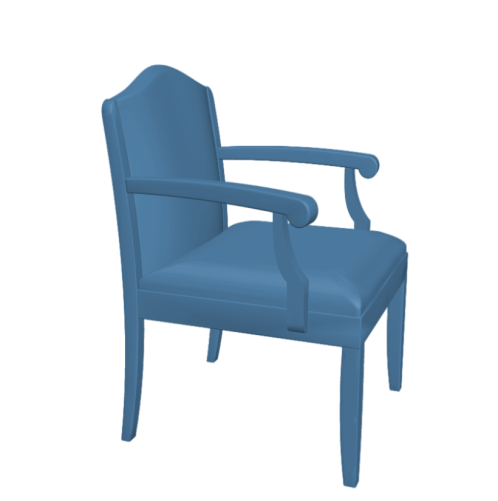}
    \caption*{}
    \end{subfigure}
\hfil
    \begin{subfigure}{0.19\linewidth}
        \includegraphics[width=\linewidth]{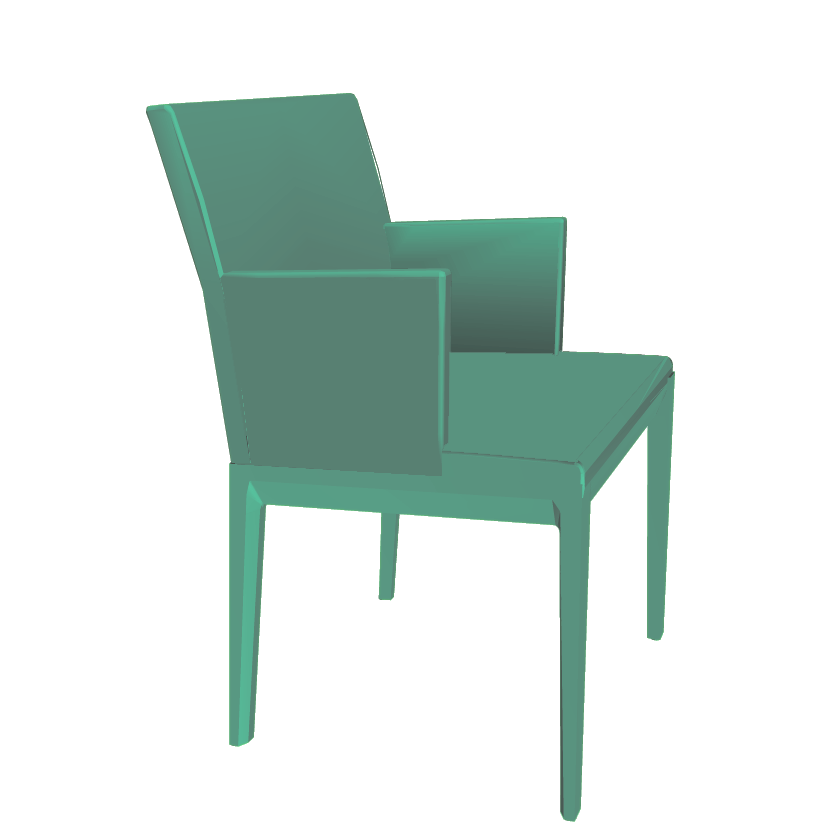}
    \caption*{}
    \end{subfigure}
\hfil
    \begin{subfigure}{0.19\linewidth}
        \includegraphics[width=\linewidth]{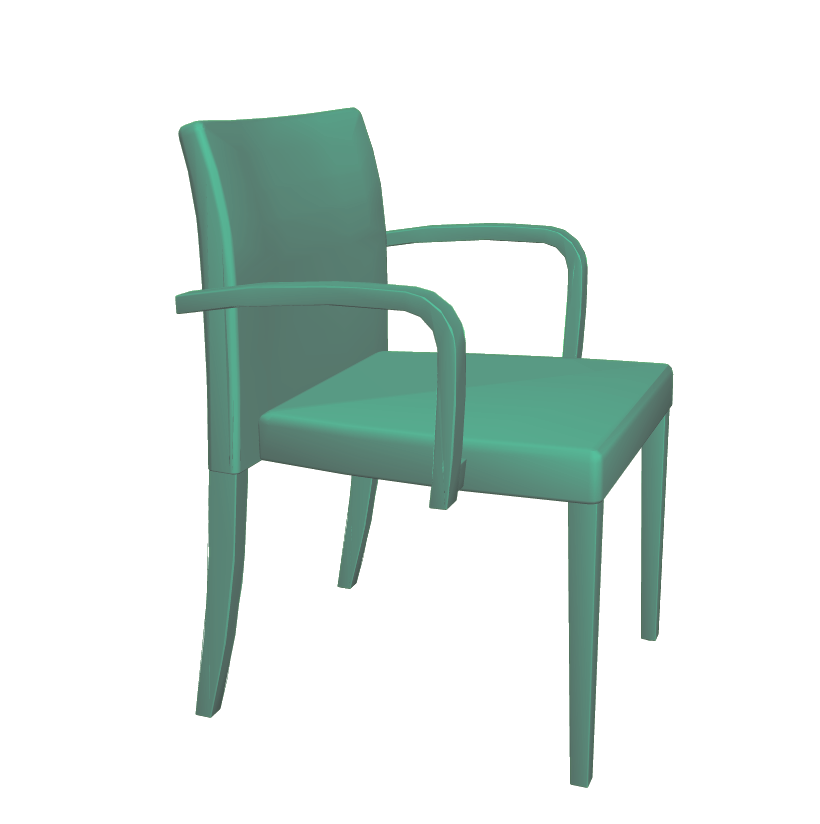}
    \caption*{}
    \end{subfigure}
\hfil
    \begin{subfigure}{0.19\linewidth}
        \includegraphics[width=\linewidth]{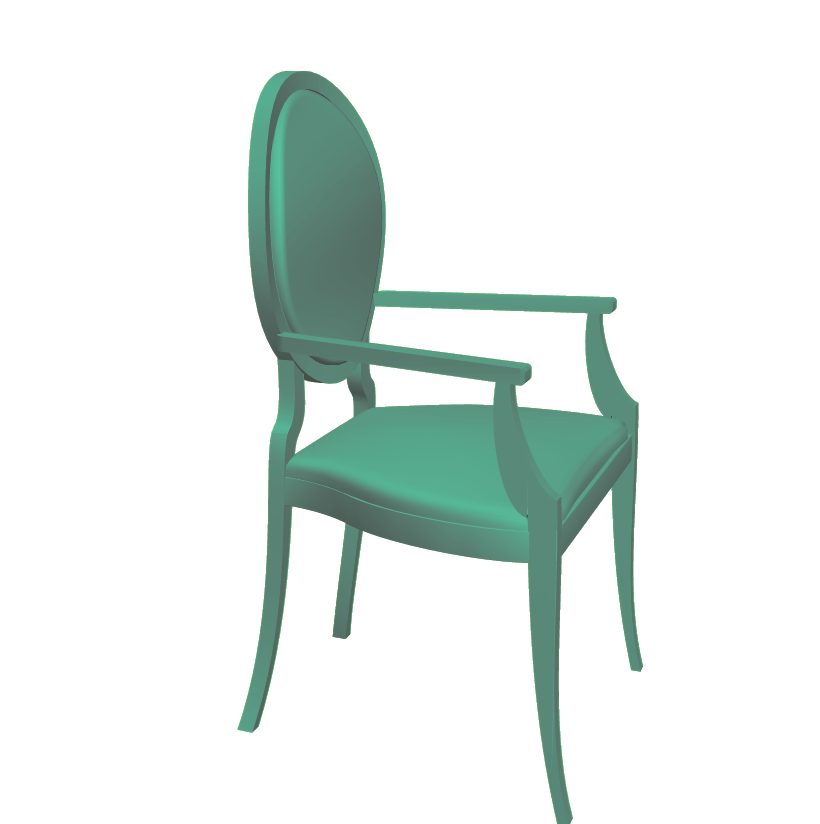}
    \caption*{}
    \end{subfigure}\\[-1.5em]
%%%%%%%%%%%%%%%%%%%%%%%%%%%%%%%%%%%%%%%%%%Chair%%%%%%%%%%%%%%%%%%%%%%%%%%%%%%%%%%%%%%%%%%%%
    \begin{subfigure}{0.19\linewidth}
        \includegraphics[width=\linewidth]{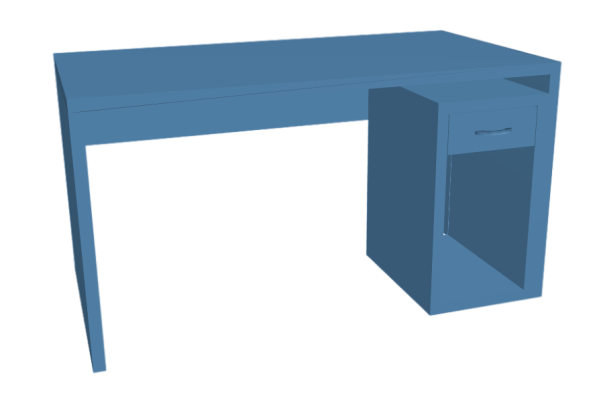}
    \caption*{}
    \end{subfigure}
\hfil
    \begin{subfigure}{0.19\linewidth}
        \includegraphics[width=\linewidth]{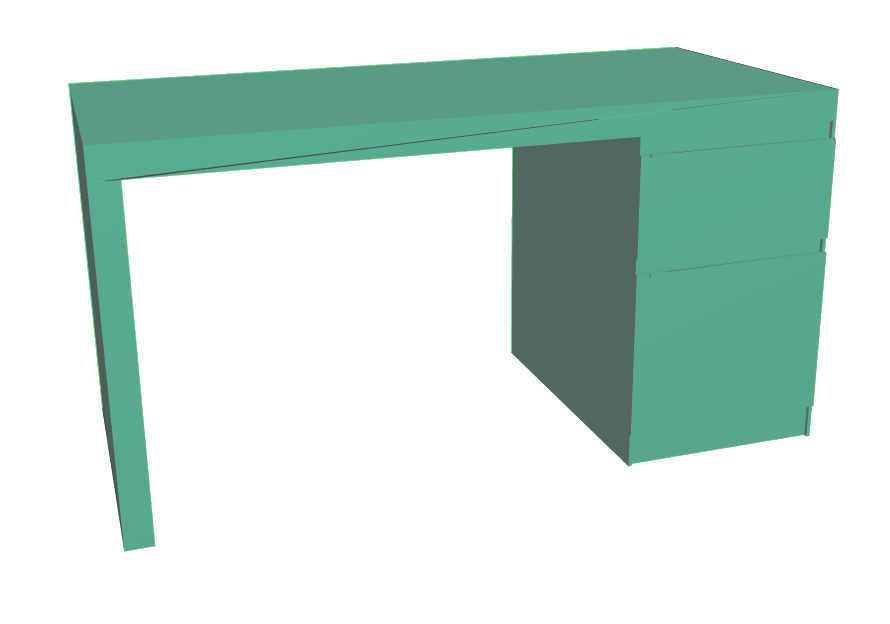}
    \caption*{}
    \end{subfigure}
\hfil
    \begin{subfigure}{0.19\linewidth}
        \includegraphics[width=\linewidth]{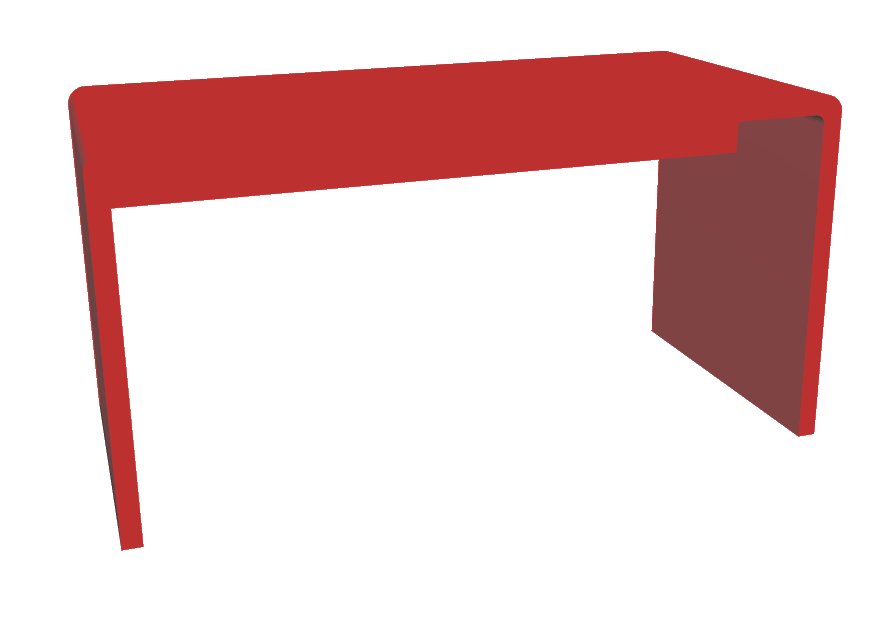}
    \caption*{}
    \end{subfigure}
\hfil
    \begin{subfigure}{0.19\linewidth}
        \includegraphics[width=\linewidth]{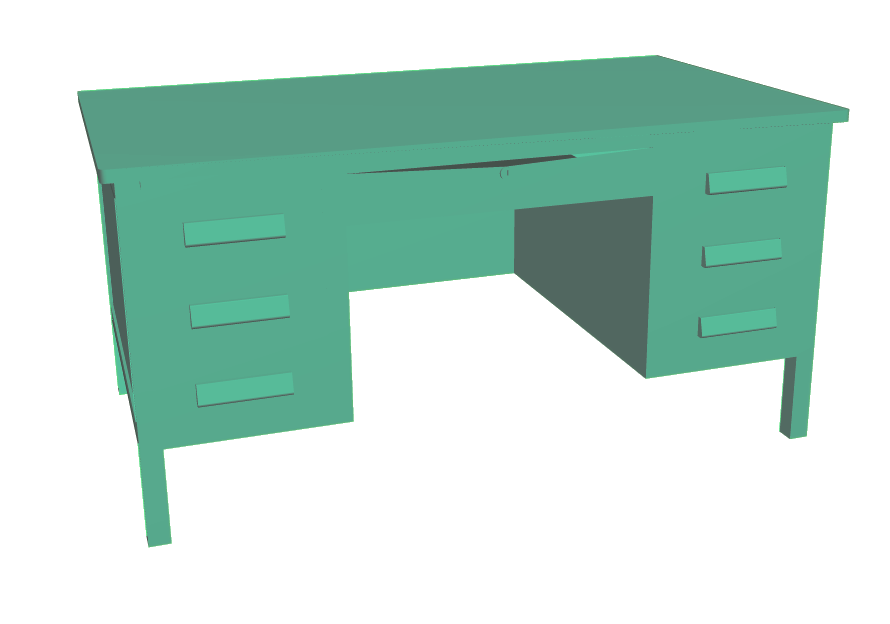}
    \caption*{}
    \end{subfigure}\\[-2em]
%%%%%%%%%%%%%%%%%%%%%%%%%%%%%%%%%%%%%%%%%%Desk%%%%%%%%%%%%%%%%%%%%%%%%%%%%%%%%%%%%%%%%%%%%
    \begin{subfigure}{0.19\linewidth}
        \includegraphics[width=\linewidth]{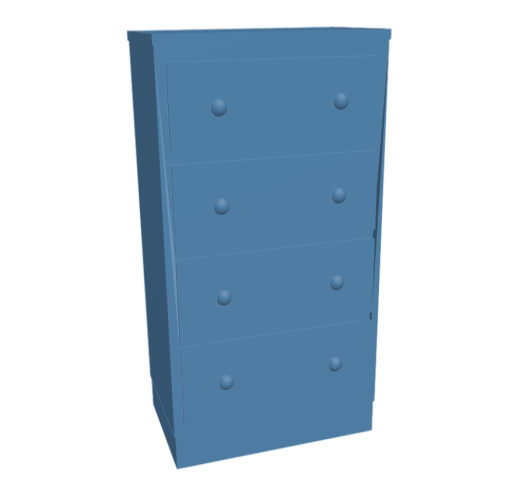}
    \caption*{}
    \end{subfigure}
\hfil
    \begin{subfigure}{0.19\linewidth}
        \includegraphics[width=\linewidth]{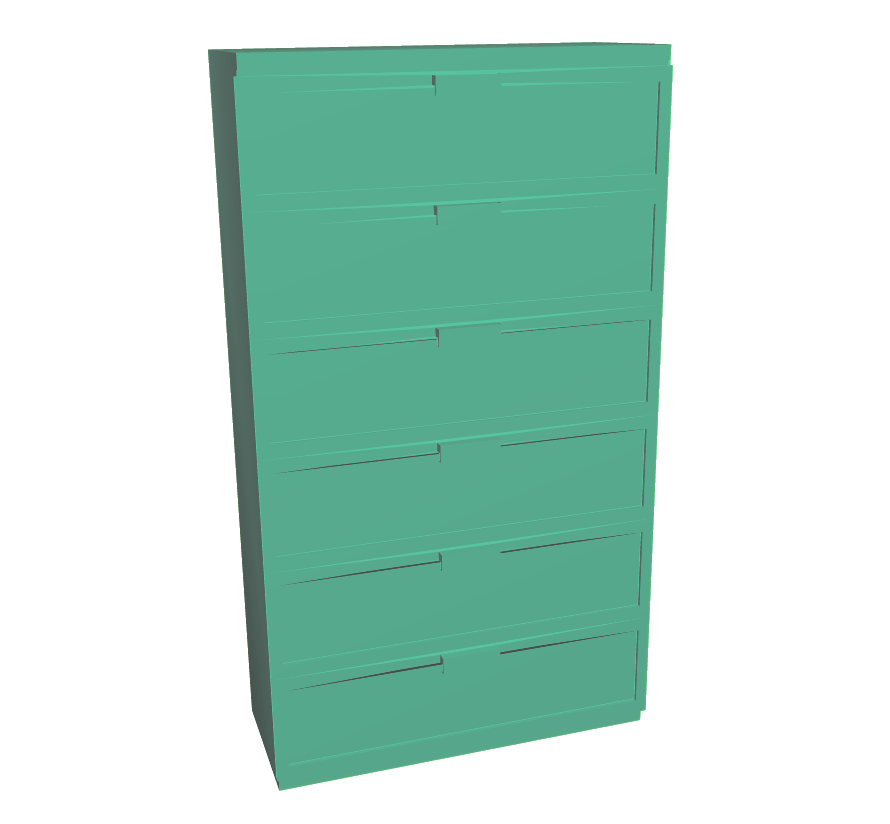}
    \caption*{}
    \end{subfigure}
\hfil
    \begin{subfigure}{0.19\linewidth}
        \includegraphics[width=\linewidth]{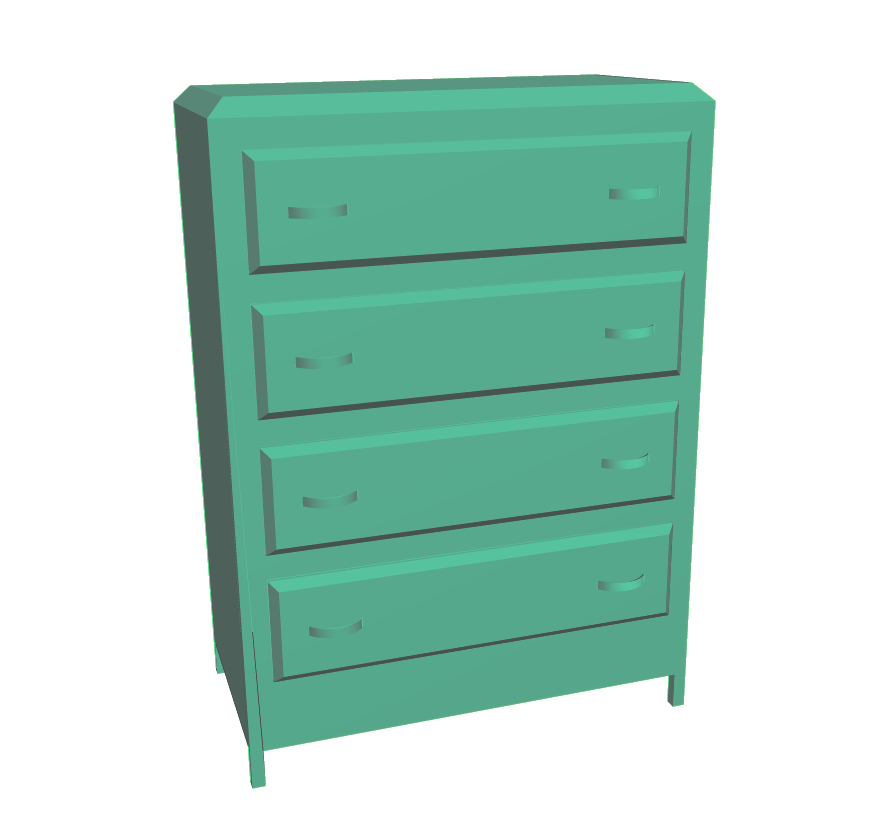}
    \caption*{}
    \end{subfigure}
\hfil
    \begin{subfigure}{0.21\linewidth}
        \includegraphics[width=\linewidth]{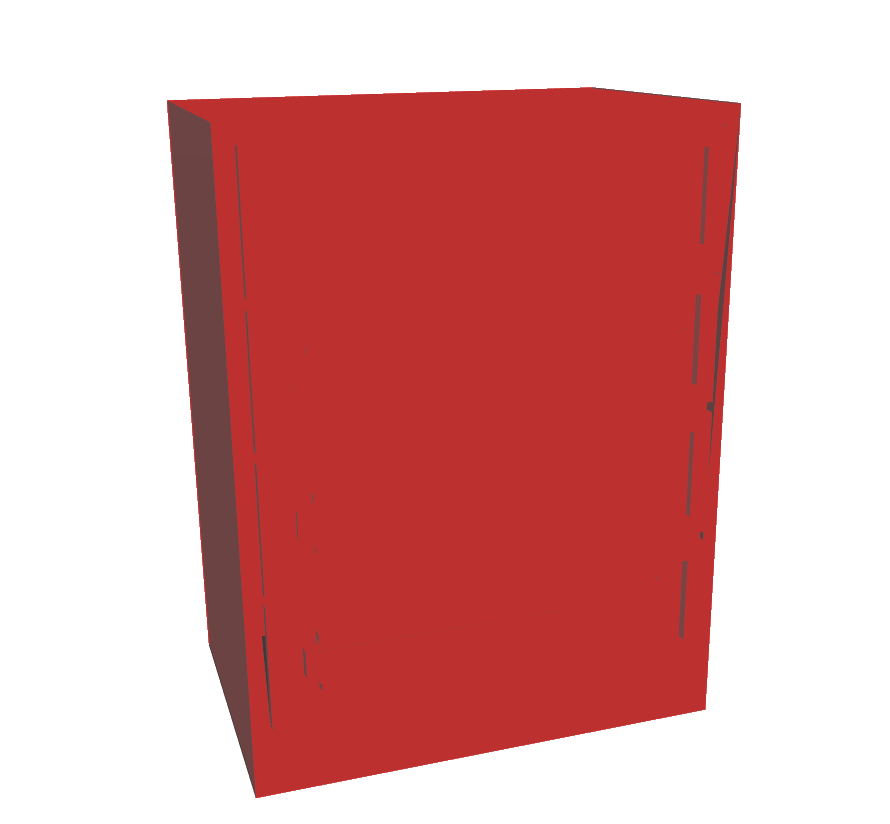}
    \caption*{}
    \end{subfigure}\\[-1.5em]
%%%%%%%%%%%%%%%%%%%%%%%%%%%%%%%%%%%%%%%%%%Dresser%%%%%%%%%%%%%%%%%%%%%%%%%%%%%%%%%%%%%%%%%%%%
    \begin{subfigure}{0.19\linewidth}
        \includegraphics[width=\linewidth]{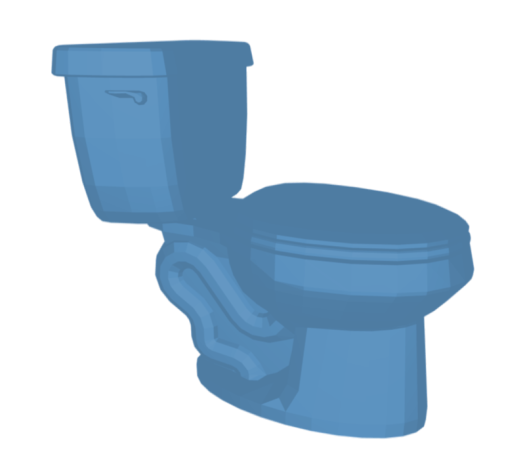}
    \caption*{}
    \end{subfigure}
\hfil
    \begin{subfigure}{0.19\linewidth}
        \includegraphics[width=\linewidth]{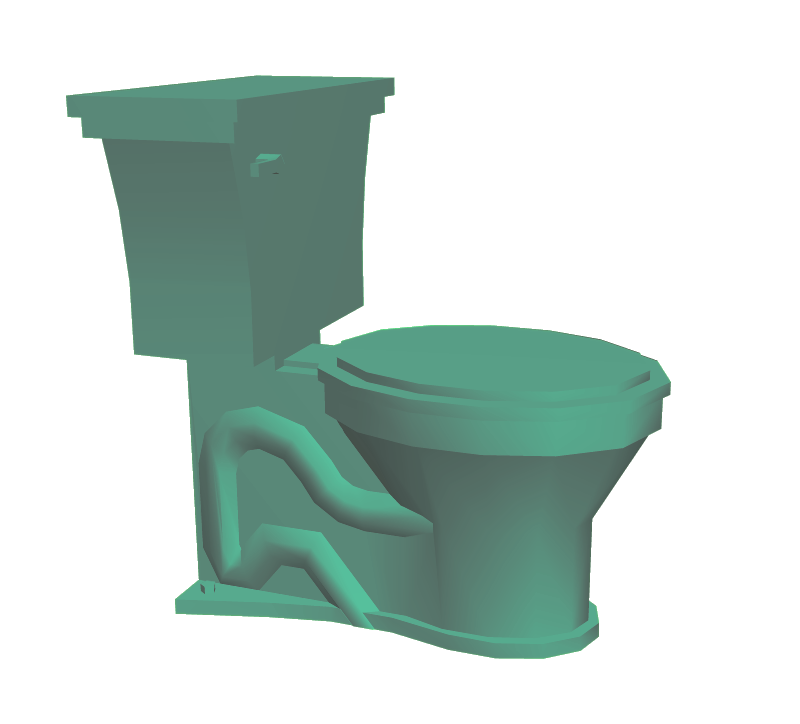}
    \caption*{}
    \end{subfigure}
\hfil
    \begin{subfigure}{0.19\linewidth}
        \includegraphics[width=\linewidth]{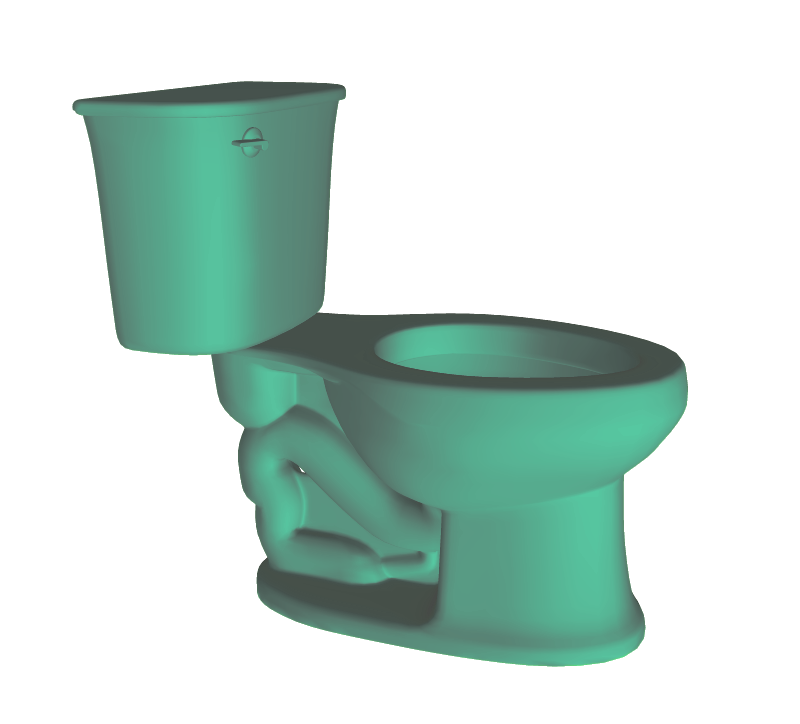}
    \caption*{}
    \end{subfigure}
\hfil
    \begin{subfigure}{0.19\linewidth}
        \includegraphics[width=\linewidth]{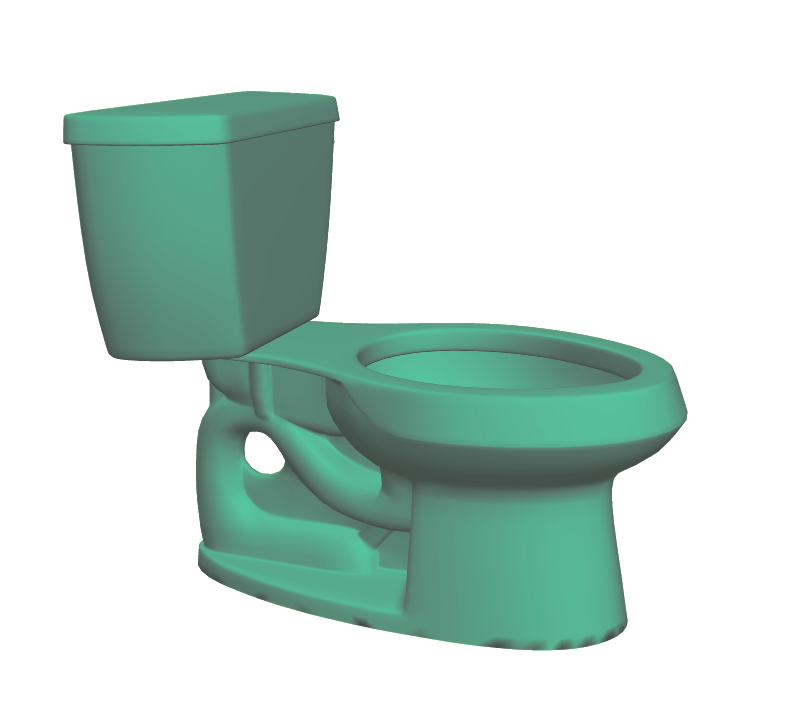}
    \caption*{}
    \end{subfigure}
    \vspace{-1em}
%%%%%%%%%%%%%%%%%%%%%%%%%%%%%%%%%%%%%%%%%%Dresser%%%%%%%%%%%%%%%%%%%%%%%%%%%%%%%%%%%%%%%%%%%%
\caption{Retrieval results. 
This figure demonstrates the retrieved shapes for the given queries using PolyNet.
The blue models in the first column are the queries.
The retrieved results in green are from the same category as the query, while the results in red are from different categories.
From left to right, the results are ordered with a descending rank.
}
    \label{fig:retrieval}
\end{figure}

%%%%%%%%%%%%%%%%%%%%%%%%%%%%%%%%%%%%%%%%%%%%%%%%%%%%%%%%%%%%%%%%%%%%%%%%%%%%%%%%%%
\subsection{3D shape retrieval}
We also evaluate and compare PolyNet in the retrieval task with previous methods.
We extract the output after the softmax, measure similarities between the query and the retrieved shapes by the L1 norm, and rank the relevant shapes.
We use mAP to quantitatively compare our retrieval approach to the related methods on both the ModelNet-10 and the ModelNet-40 datasets, as shown in Table~\ref{table:sota}.
We outperform all previously evaluated methods on the retrieval task based on polygon mesh and voxel grid representations.
Lastly, we show some retrieved shapes in a ranked order for the given queries on the ModelNet-10 trained by squeezed PolyConvs, including polynomial functions of degree $d=2$ in Figure~\ref{fig:retrieval}. We illustrate that our method can retrieve visually similar shapes even when the query and retrieved shapes are in different categories (e.g., retrieved table for the query desk and nightstand for the dresser).

\section{Conclusion}
In this paper, we propose PolyNet, a DNN-based method consists of PolyConv and PolyPool operations to locally learn and aggregate the information on the surface of 3D shapes.
In PolyConv, we utilize polynomial functions to learn a continuous distribution as the convolutional filters, which is 
invariant to the variation in the degree of vertices, their permutations, and their pairwise distances. 
Moreover, we design PolyShape with a multi-resolution structure that enables applying PolyPool operation without missing geometrical structures after each layer. 
Our comprehensive evaluations of PolyNet across classification and retrieval tasks and the theoretical analysis indicating the invariant properties of PolyConv demonstrate its strength and superiority over most of the previous methods.
In future works, we will explore the applications of PolyNet in 3D shape segmentation and  PolyConv in image-based computer vision tasks where there are no regular neighboring connectives.

\section*{Acknowledgement}
This work was supported by IITP grant funded by the Korea government(MSIT) [NO.2021-0-01343, Artificial Intelligence Graduate School Program (Seoul National University)]

{\small
\bibliographystyle{ieee_fullname}
\bibliography{polynet}
}

\threedvfinalcopy % *** Uncomment this line for the final submission

\def\threedvPaperID{349} % *** Enter the 3DV Paper ID here
\def\httilde{\mbox{\tt\raisebox{-.5ex}{\symbol{126}}}}

% Pages are numbered in submission mode, and unnumbered in camera-ready
\ifthreedvfinal\pagestyle{empty}\fi

%%%%%%%%% TITLE
\title{\bf Supplementary Material \textit{for} \\ \vspace{2mm} PolyNet: Polynomial Neural Network for 3D Shape Recognition with PolyShape Representation}

\author{
Mohsen Yavartanoo$^{1}$\quad Shih-Hsuan Hung$^{2}$\quad Reyhaneh Neshatavar$^{1}$\quad Yue Zhang$^{2}$\quad Kyoung Mu Lee$^{1}$\\
\hspace{1.5cm}$^{1}$SNU ECE \& ASRI \hspace{4.75cm}
$^{2}$Oregon State University\\{\small \texttt {\{myavartanoo,reyhanehneshat,kyoungmu\}@snu.ac.kr} \hspace{1.0cm} \texttt{\{hungsh,zhangyue\}@oregonstate.edu}}
}


\maketitle
\ifthreedvfinal\thispagestyle{empty}\fi

\appendixpageoff
\appendixtitleoff
\begin{appendices}
  \setcounter{section}{18}

\section{PolyNet analysis}
In this section, we provide more mathematical and qualitative analysis of our proposed PolyNet with additional explanations of PolyConv operation and the preprocessing procedure of PolyShape and its results. 
First, we analyze some equations of PolyConv mentioned in the main paper for better understanding. 
Then we discuss PolyShape preprocessing in detail and show its step-by-step results. 

\subsection{Expansion of PolyConv}\label{expansion}
%To clear the paramertrization of the coefficent matrix $A$ of 
We derive the compact form of the polynomial function $f(x,y)$ defined in Eq.~\ref{implicit_5} in the  main paper for $d=2$ and $d=4$ as in the following Eq.~\ref{eq1} and Eq.~\ref{eq2}, respectively. 

\begin{equation}\tag{S1}
\begin{split}
    & X^TAX \\
    & = \begin{bmatrix}
        1 & x & y\\
    \end{bmatrix}
    \begin{bmatrix}
        A_{11} & A_{12} & A_{13}\\
        A_{21} & A_{22} & A_{23}\\
        A_{31} & A_{32} & A_{33}
    \end{bmatrix}
    \begin{bmatrix}
        1\\
        x\\
        y
    \end{bmatrix} \\
     &= A_{11}\\
     &+(A_{12}+A_{21})x +(A_{13}+A_{31})y \\
     &+A_{22}x^2 +(A_{23}+A_{32})xy+A_{33}y^2 \\
     &= a_{0,0} + a_{1,0}x + a_{0,1}y + a_{2,0}x^2 + a_{1,1}xy\\
     &+ a_{0,2}y^2 \\
     &= \sum_{0\leq i,j,i+j\leq 2} a_{i,j}x^iy^j = f(x,y).
\end{split}
\label{eq1}
\end{equation}

\begin{equation}\tag{S2}
\begin{split}
    & X^TAX\\
    & = \begin{bmatrix}
        1 & x & y & x^2 & xy & y^2\\
    \end{bmatrix}\\
    &\begin{bmatrix}
        A_{11} & A_{12} & A_{13} & A_{14} & A_{15} & A_{16}\\
        A_{21} & A_{22} & A_{23} & A_{24} & A_{25} & A_{26}\\
        A_{31} & A_{32} & A_{33} & A_{34} & A_{35} & A_{36}
    \end{bmatrix}
    \begin{bmatrix}
        1\\
        x\\
        y\\
        x^2\\
        xy\\
        y^2
    \end{bmatrix} \\
     &= A_{11}\\
     &+(A_{12}+A_{21})x +(A_{13}+A_{31})y\\
     &+(A_{22}+A_{15}
     +A_{51})x^2 +(A_{23}+A_{32}\\
     &+A_{14}+A_{41})xy+(A_{33}+A_{16}+A_{61})y^2\\
     &+(A_{25}+A_{52})x^3+(A_{35}+A_{53}+A_{24}\\
     &+A_{42})x^2y+(A_{26}+A_{62}+A_{34}+A_{43})xy^2\\
     &+(A_{36}+A_{63})y^3\\
     &+A_{55}x^4+(A_{45}+A_{54})x^3y+(A_{44}+A_{56}\\
     &+A_{65})x^2y^2+(A_{46}+A_{64})xy^3+A_{66}y^4 \\
     &= a_{0,0} + a_{1,0}x + a_{0,1}y\\
     &+ a_{2,0}x^2 + a_{1,1}xy + a_{0,2}y^2\\
     &+a_{3,0}x^3+a_{2,1}x^2y+a_{1,2}xy^2+a_{0,3}y^3\\
     &+a_{4,0}x^4+a_{3,1}x^3y+a_{2,2}x^2y^2+a_{1,3}xy^3\\
     &+a_{0,4}y^4 \\
     &= \sum_{0\leq i,j,i+j\leq 4} a_{i,j}x^iy^j = f(x,y).
\end{split}
\label{eq2}
\end{equation}

\renewcommand{\thefigure}{S1}
\begin{figure*}[h]
\begin{center}
\includegraphics[width=1.0\textwidth]{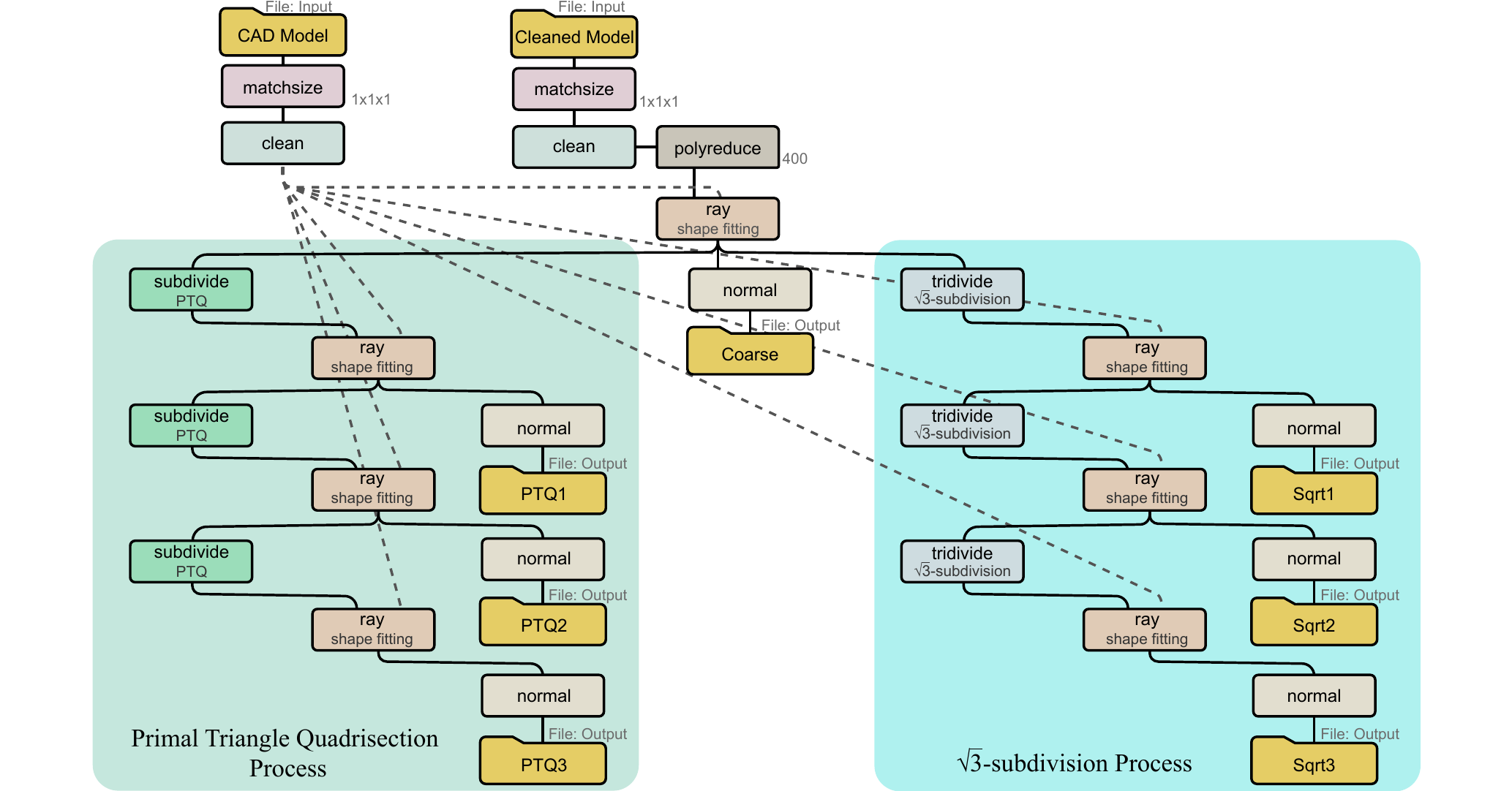}
\end{center}
\caption{PolyShape processing in Houdini. Each box refers to a node with the same name in the software.}
\label{fig:houdini}
\end{figure*}

Note that we parameterize the matrix $A$ by the matrix $B$ and learn the matrix $B$ instead of the matrix $A$. The parametrized matrix $A$ for the polynomial functions of degree $d=2$ is as Eq.~\ref{eq3};
\begin{equation}\tag{S3}
\begin{split}
A &= BB^T \\
   & =\begin{bmatrix}
        A_{11} & A_{12} & A_{13}\\
        A_{21} & A_{22} & A_{23}\\
        A_{31} & A_{32} & A_{33}
    \end{bmatrix}\\
    & =\begin{bmatrix}
        B_{11} & B_{12} & B_{13}\\
        B_{12} & B_{22} & B_{23}\\
        B_{13} & B_{23} & B_{33}
    \end{bmatrix}
    \begin{bmatrix}
        B_{11} & B_{12} & B_{13}\\
        B_{12} & B_{22} & B_{23}\\
        B_{13} & B_{23} & B_{33}
    \end{bmatrix}\\
    &\Longrightarrow A_{11} = B_{11}^2+B_{12}^2+B_{13}^2\\
    &\Longrightarrow A_{12} = B_{11}B_{12}+B_{12}B_{22}+B_{13}B_{23}\\
    &\Longrightarrow A_{13} = B_{11}B_{13}+B_{12}B_{23}+B_{13}B_{33}\\
    &\Longrightarrow A_{21} = B_{12}B_{11}+B_{22}B_{12}+B_{23}B_{13}\\
    &\Longrightarrow A_{22} = B_{12}^2+B_{22}^2+B_{23}^2\\
    &\Longrightarrow A_{23} = B_{12}B_{13}+B_{22}B_{23}+B_{23}B_{33}\\
    &\Longrightarrow A_{31} = B_{13}B_{11}+B_{23}B_{12}+B_{33}B_{13}\\
    &\Longrightarrow A_{32} = B_{13}B_{12}+B_{23}B_{22}+B_{33}B_{23}\\
    &\Longrightarrow A_{33} = B_{13}^2+B_{23}^2+B_{33}^2,
\end{split}
\label{eq3}
\end{equation}
where $B$ is a learnable symmetric matrix.

\subsection{Details of PolyShape Processing}
We provide the flowchart of the PolyShape processing in Houdini software in Figure~\ref{fig:houdini}. 
Given a 3D CAD model and its cleaned model by mesh fusion, we first resize the models and remove the unused points with the \textit{matchsize} and \textit{clean} nodes.
Next, we generate the coarse mesh by reducing the number of vertices to $400$ with \textit{polyreduce} and fitting the shape to the 3D CAD model with \textit{ray}.
We create the multiresolution of the PolyShape by subdividing the coarse mesh three times with primal triangle quadrisection (PTQ, \textit{subdivide}) or $\sqrt{3}$-subdivision (\textit{tridivide}), respectively.  
At each iteration of these subdivisions, we fit the generated mesh to the given 3D CAD model to maintain the details of the original shape.
Figure~\ref{fig:SQRT} and Figure~\ref{fig:PTQ} show the resulting PolyShapes of $\sqrt{3}$-subdivision and PTQ with the same input models, respectively.
The PolyShapes generated by $\sqrt{3}$-subdivision have fewer faces than the shapes created by PTQ at the same level of details.
For the highest resolution of the PolyShapes, $\sqrt{3}$-subdivision creates $0.43$ fewer faces than PTQ on average.
Therefore, the $\sqrt{3}$-subdivision provides a more efficient representation for storage and the computation of the classification.

\renewcommand{\thefigure}{S2}
\begin{figure*}
\captionsetup[subfigure]{justification=centering}
\centering
    \begin{subfigure}{0.10\linewidth}
        \includegraphics[width=\linewidth]{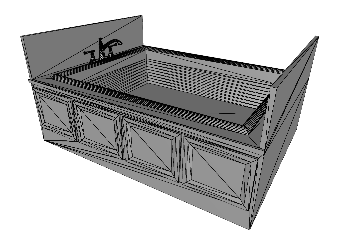}
    \caption*{\tiny $V=15398$ \\$F=19254$}
    \end{subfigure}
\hfil
    \begin{subfigure}{0.10\linewidth}
        \includegraphics[width=\linewidth]{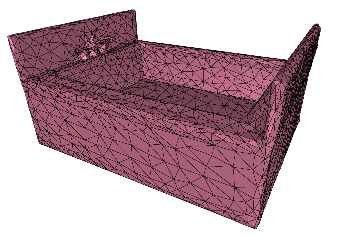}
    \caption*{\tiny$V=2500$ \\ \tiny$F=4996$}
    \end{subfigure}
\hfil
    \begin{subfigure}{0.10\linewidth}
        \includegraphics[width=\linewidth]{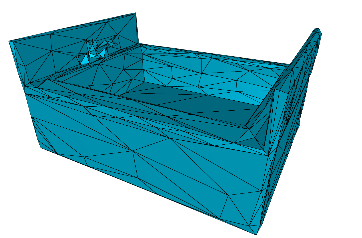}
    \caption*{\tiny$V=400$ \\ \tiny$F=796$}
    \end{subfigure}
\hfil
    \begin{subfigure}{0.10\linewidth}
        \includegraphics[width=\linewidth]{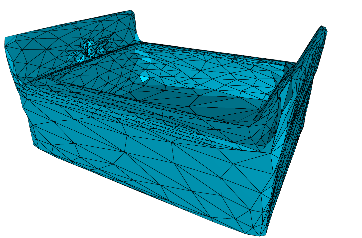}
    \caption*{\tiny$V=1198$ \\ \tiny$F=2394$}
    \end{subfigure}
\hfil
    \begin{subfigure}{0.10\linewidth}
        \includegraphics[width=\linewidth]{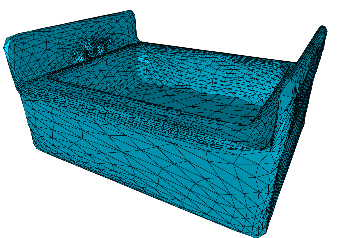}
    \caption*{\tiny$V=3592$ \\ \tiny$F=7182$}
    \end{subfigure}
\hfil
    \begin{subfigure}{0.10\linewidth}
        \includegraphics[width=\linewidth]{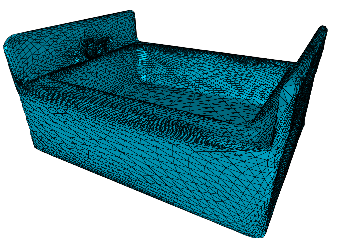}
    \caption*{\tiny$V=10774$ \\ \tiny$F=21546$}
    \end{subfigure}
%%%%%%%%%%%%%%%%%%%%%%%%%%%%%%%%%%%%%%%%%%%%%%%%%%%%%%%%%%%%%%%%%%%%

    \begin{subfigure}{0.10\linewidth}
        \includegraphics[width=\linewidth]{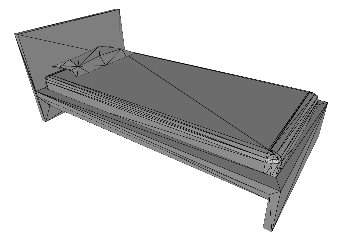}
    \caption*{\tiny$V=584$ \\ \tiny$F=676$}
    \end{subfigure}
\hfil
    \begin{subfigure}{0.10\linewidth}
        \includegraphics[width=\linewidth]{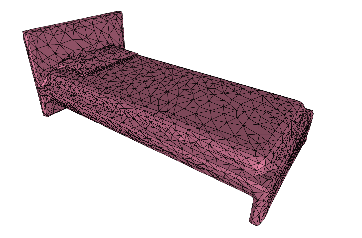}
    \caption*{\tiny$V=2502$ \\ \tiny$F=5000$}
    \end{subfigure}
\hfil
    \begin{subfigure}{0.10\linewidth}
        \includegraphics[width=\linewidth]{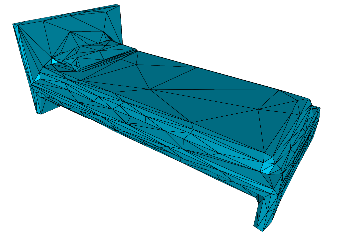}
    \caption*{\tiny$V=400$ \\ \tiny$F=796$}
    \end{subfigure}
\hfil
    \begin{subfigure}{0.10\linewidth}
        \includegraphics[width=\linewidth]{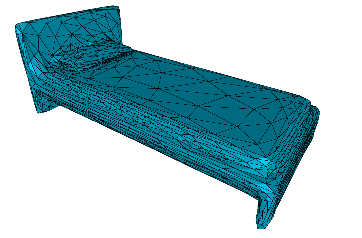}
    \caption*{\tiny$V=1196$ \\ \tiny$F=2388$}
    \end{subfigure}
\hfil
    \begin{subfigure}{0.10\linewidth}
        \includegraphics[width=\linewidth]{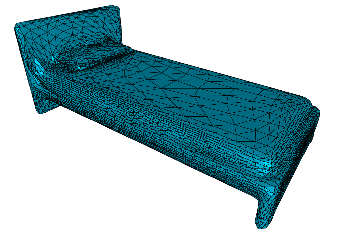}
    \caption*{\tiny$V=3584$ \\ \tiny$F=7164$}
    \end{subfigure}
\hfil
    \begin{subfigure}{0.10\linewidth}
        \includegraphics[width=\linewidth]{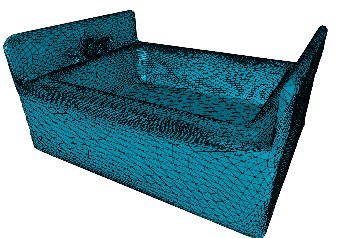}
    \caption*{\tiny$V=10748$ \\ \tiny$F=21492$}
    \end{subfigure}
%%%%%%%%%%%%%%%%%%%%%%%%%%%%%%%%%%%%%%%%%%%%%%%%%%%%%%%%%%%%%%%%%%%%

    \begin{subfigure}{0.10\linewidth}
        \includegraphics[width=\linewidth]{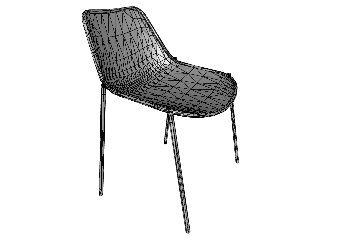}
    \caption*{\tiny$V=584$ \\ \tiny$F=676$}
    \end{subfigure}
\hfil
    \begin{subfigure}{0.10\linewidth}
        \includegraphics[width=\linewidth]{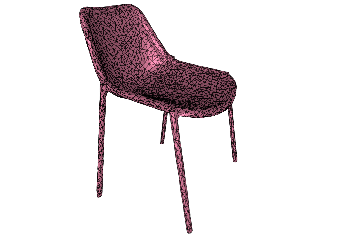}
    \caption*{\tiny$V=2502$ \\ \tiny$F=5000$}
    \end{subfigure}
\hfil
    \begin{subfigure}{0.10\linewidth}
        \includegraphics[width=\linewidth]{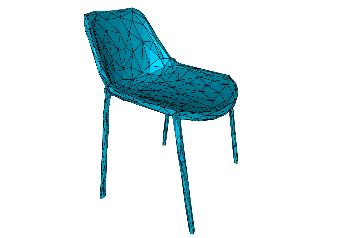}
    \caption*{\tiny$V=400$ \\ \tiny$F=796$}
    \end{subfigure}
\hfil
    \begin{subfigure}{0.10\linewidth}
        \includegraphics[width=\linewidth]{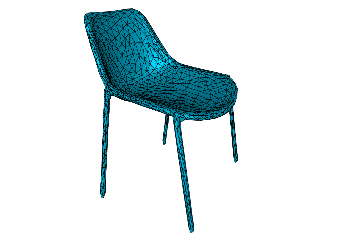}
    \caption*{\tiny$V=1196$ \\ \tiny$F=2388$}
    \end{subfigure}
\hfil
    \begin{subfigure}{0.10\linewidth}
        \includegraphics[width=\linewidth]{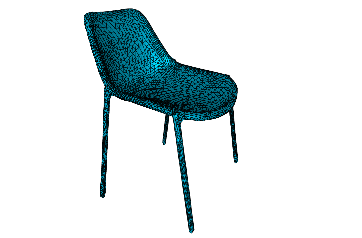}
    \caption*{\tiny$V=3584$ \\ \tiny$F=7164$}
    \end{subfigure}
\hfil
    \begin{subfigure}{0.10\linewidth}
        \includegraphics[width=\linewidth]{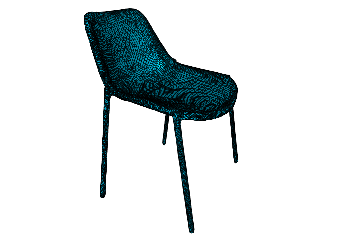}
    \caption*{\tiny$V=10748$ \\ \tiny$F=21492$}
    \end{subfigure}
%%%%%%%%%%%%%%%%%%%%%%%%%%%%%%%%%%%%%%%%%%%%%%%%%%%%%%%%%%%%%%%%%%%%

    \begin{subfigure}{0.10\linewidth}
        \includegraphics[width=\linewidth]{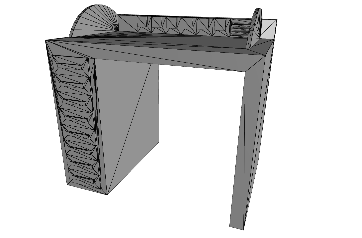}
    \caption*{\tiny$V=1988$ \\ \tiny$F=1376$}
    \end{subfigure}
\hfil
    \begin{subfigure}{0.10\linewidth}
        \includegraphics[width=\linewidth]{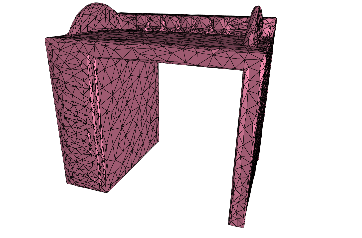}
    \caption*{\tiny$V=2273$ \\ \tiny$F=4542$}
    \end{subfigure}
\hfil
    \begin{subfigure}{0.10\linewidth}
        \includegraphics[width=\linewidth]{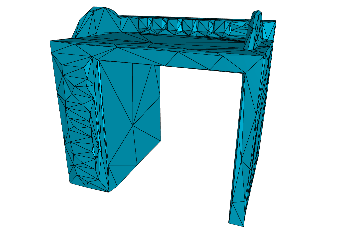}
    \caption*{\tiny$V=400$ \\ \tiny$F=816$}
    \end{subfigure}
\hfil
    \begin{subfigure}{0.10\linewidth}
        \includegraphics[width=\linewidth]{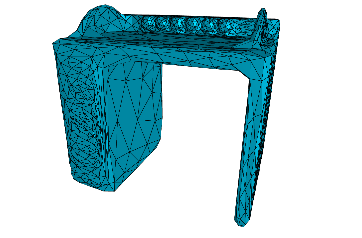}
    \caption*{\tiny$V=1216$ \\ \tiny$F=2448$}
    \end{subfigure}
\hfil
    \begin{subfigure}{0.10\linewidth}
        \includegraphics[width=\linewidth]{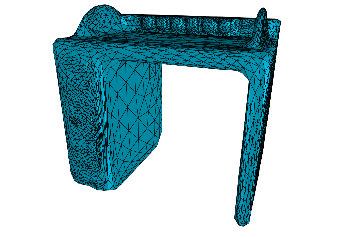}
    \caption*{\tiny$V=3664$ \\ \tiny$F=7344$}
    \end{subfigure}
\hfil
    \begin{subfigure}{0.10\linewidth}
        \includegraphics[width=\linewidth]{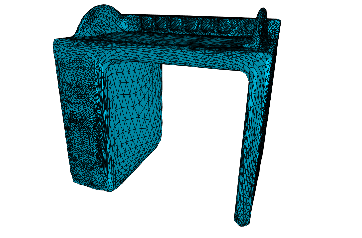}
    \caption*{\tiny$V=11008$ \\ \tiny$F=22032$}
    \end{subfigure}
%%%%%%%%%%%%%%%%%%%%%%%%%%%%%%%%%%%%%%%%%%%%%%%%%%%%%%%%%%%%%%%%%%%%

    \begin{subfigure}{0.10\linewidth}
        \includegraphics[width=\linewidth]{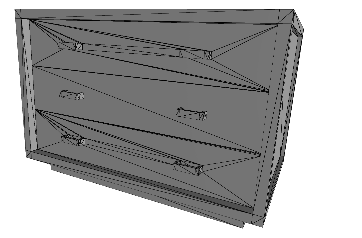}
            \caption*{\tiny$V=3958$ \\ \tiny$F=3132$}
    \end{subfigure}
\hfil
    \begin{subfigure}{0.10\linewidth}
        \includegraphics[width=\linewidth]{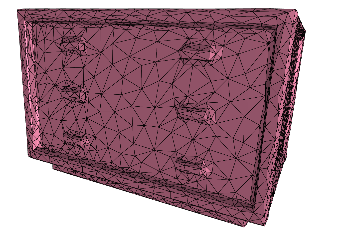}
    \caption*{\tiny$V=2502$ \\ \tiny$F=5000$}
    \end{subfigure}
\hfil
    \begin{subfigure}{0.10\linewidth}
        \includegraphics[width=\linewidth]{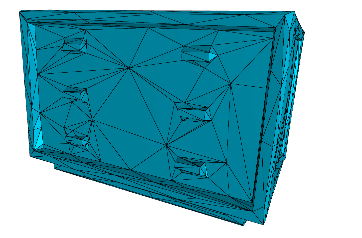}
    \caption*{\tiny$V=400$ \\ \tiny$F=796$}
    \end{subfigure}
\hfil
    \begin{subfigure}{0.10\linewidth}
        \includegraphics[width=\linewidth]{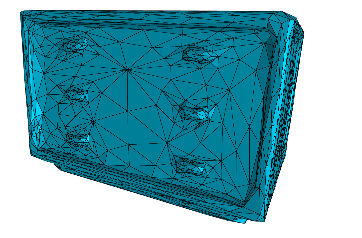}
    \caption*{\tiny$V=1196$ \\ \tiny$F=2388$}
    \end{subfigure}
\hfil
    \begin{subfigure}{0.10\linewidth}
        \includegraphics[width=\linewidth]{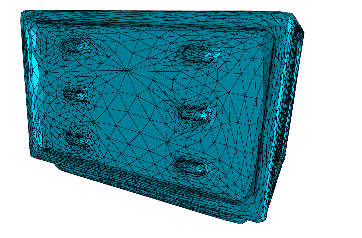}
    \caption*{\tiny$V=3584$ \\ \tiny$F=7164$}
    \end{subfigure}
\hfil
    \begin{subfigure}{0.10\linewidth}
        \includegraphics[width=\linewidth]{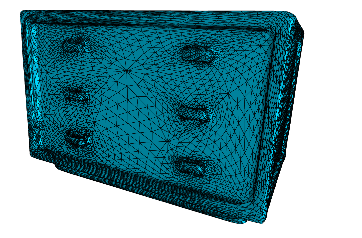}
    \caption*{\tiny$V=10748$ \\ \tiny$F=21492$}
    \end{subfigure}
%%%%%%%%%%%%%%%%%%%%%%%%%%%%%%%%%%%%%%%%%%%%%%%%%%%%%%%%%%%%%%%%%%%%

    \begin{subfigure}{0.10\linewidth}
        \includegraphics[width=\linewidth]{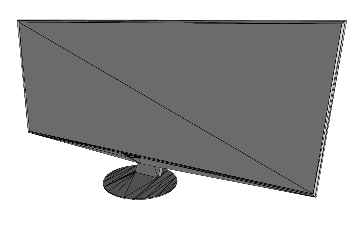}
            \caption*{\tiny$V=2482$ \\ \tiny$F=1794$}
    \end{subfigure}
\hfil
    \begin{subfigure}{0.10\linewidth}
        \includegraphics[width=\linewidth]{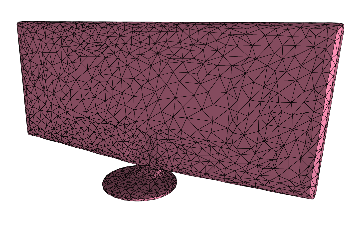}
            \caption*{\tiny$V=2500$ \\ \tiny$F=5000$}
    \end{subfigure}
\hfil
    \begin{subfigure}{0.10\linewidth}
        \includegraphics[width=\linewidth]{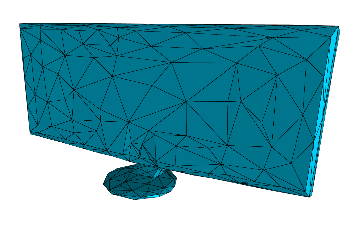}
            \caption*{\tiny$V=400$ \\ \tiny$F=800$}
    \end{subfigure}
\hfil
    \begin{subfigure}{0.10\linewidth}
        \includegraphics[width=\linewidth]{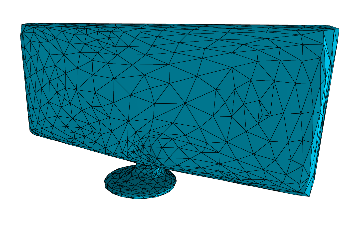}
            \caption*{\tiny$V=1200$ \\ \tiny$F=2400$}
    \end{subfigure}
\hfil
    \begin{subfigure}{0.10\linewidth}
        \includegraphics[width=\linewidth]{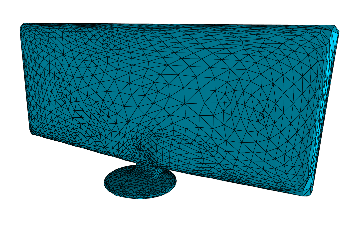}
            \caption*{\tiny$V=3600$ \\ \tiny$F=7200$}
    \end{subfigure}
\hfil
    \begin{subfigure}{0.10\linewidth}
        \includegraphics[width=\linewidth]{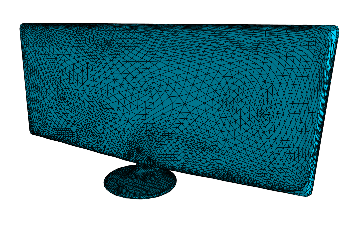}
            \caption*{\tiny$V=10800$ \\ \tiny$F=21600$}
    \end{subfigure}
%%%%%%%%%%%%%%%%%%%%%%%%%%%%%%%%%%%%%%%%%%%%%%%%%%%%%%%%%%%%%%%%%%%%

    \begin{subfigure}{0.10\linewidth}
        \includegraphics[width=\linewidth]{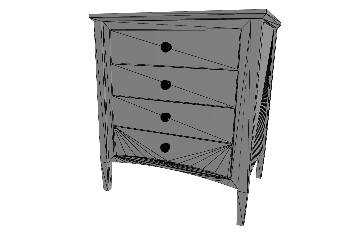}
            \caption*{\tiny$V=41528$ \\ \tiny$F=22000$}
    \end{subfigure}
\hfil
    \begin{subfigure}{0.10\linewidth}
        \includegraphics[width=\linewidth]{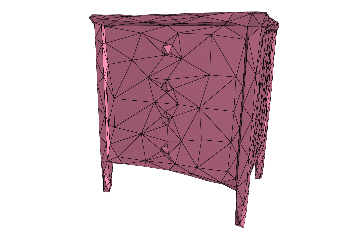}
            \caption*{\tiny$V=2037$ \\ \tiny$F=4074$}
    \end{subfigure}
\hfil
    \begin{subfigure}{0.10\linewidth}
        \includegraphics[width=\linewidth]{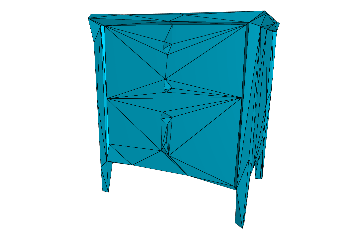}
            \caption*{\tiny$V=400$ \\ \tiny$F=942$}
    \end{subfigure}
\hfil
    \begin{subfigure}{0.10\linewidth}
        \includegraphics[width=\linewidth]{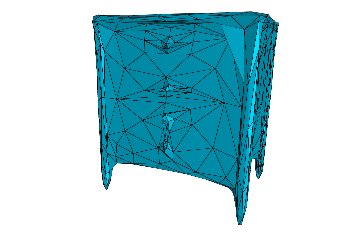}
            \caption*{\tiny$V=1342$ \\ \tiny$F=2826$}
    \end{subfigure}
\hfil
    \begin{subfigure}{0.10\linewidth}
        \includegraphics[width=\linewidth]{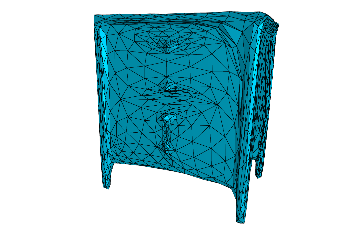}
            \caption*{\tiny$V=4168$ \\ \tiny$F=8478$}
    \end{subfigure}
\hfil
    \begin{subfigure}{0.10\linewidth}
        \includegraphics[width=\linewidth]{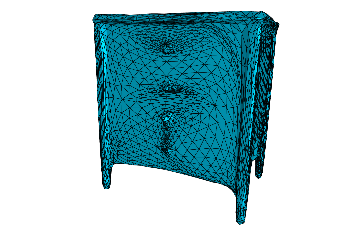}
            \caption*{\tiny$V=12646$ \\ \tiny$F=25434$}
    \end{subfigure}
%%%%%%%%%%%%%%%%%%%%%%%%%%%%%%%%%%%%%%%%%%%%%%%%%%%%%%%%%%%%%%%%%%%%

    \begin{subfigure}{0.10\linewidth}
        \includegraphics[width=\linewidth]{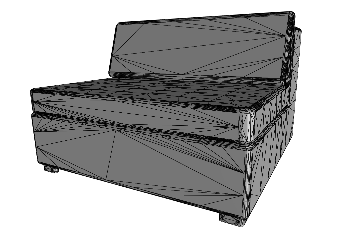}
            \caption*{\tiny$V=7808$ \\ \tiny$F=13980$}
    \end{subfigure}
\hfil
    \begin{subfigure}{0.10\linewidth}
        \includegraphics[width=\linewidth]{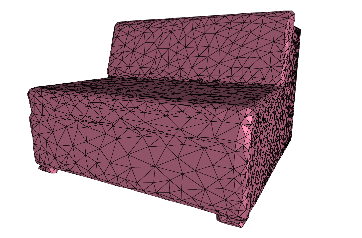}
            \caption*{\tiny$V=2496$ \\ \tiny$F=4988$}
    \end{subfigure}
\hfil
    \begin{subfigure}{0.10\linewidth}
        \includegraphics[width=\linewidth]{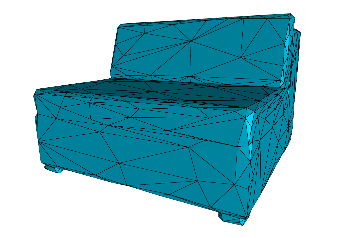}
            \caption*{\tiny$V=400$ \\ \tiny$F=798$}
    \end{subfigure}
\hfil
    \begin{subfigure}{0.10\linewidth}
        \includegraphics[width=\linewidth]{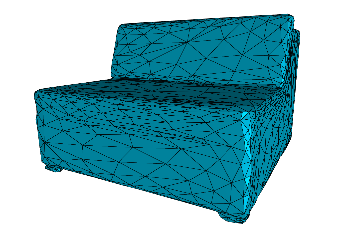}
            \caption*{\tiny$V=1198$ \\ \tiny$F=2394$}
    \end{subfigure}
\hfil
    \begin{subfigure}{0.10\linewidth}
        \includegraphics[width=\linewidth]{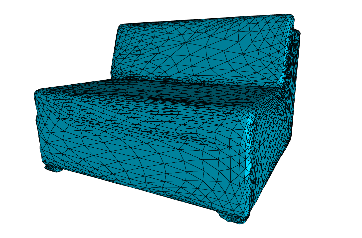}
            \caption*{\tiny$V=3592$ \\ \tiny$F=7182$}
    \end{subfigure}
\hfil
    \begin{subfigure}{0.10\linewidth}
        \includegraphics[width=\linewidth]{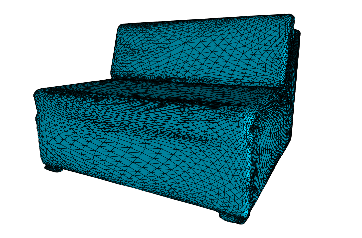}
            \caption*{\tiny$V=10774$ \\ \tiny$F=21546$}
    \end{subfigure}
%%%%%%%%%%%%%%%%%%%%%%%%%%%%%%%%%%%%%%%%%%%%%%%%%%%%%%%%%%%%%%%%%%%%

    \begin{subfigure}{0.10\linewidth}
        \includegraphics[width=\linewidth]{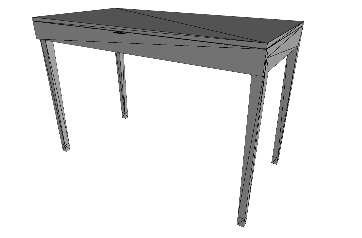}
            \caption*{\tiny$V=1074$ \\ \tiny$F=694$}
    \end{subfigure}
\hfil
    \begin{subfigure}{0.10\linewidth}
        \includegraphics[width=\linewidth]{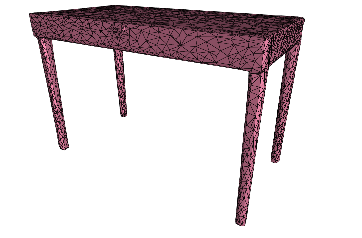}
            \caption*{\tiny$V=2502$ \\ \tiny$F=5000$}
    \end{subfigure}
\hfil
    \begin{subfigure}{0.10\linewidth}
        \includegraphics[width=\linewidth]{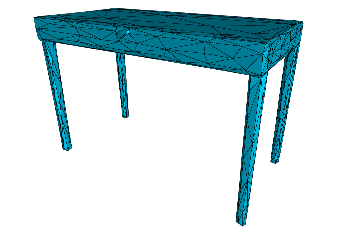}
            \caption*{\tiny$V=400$ \\ \tiny$F=796$}
    \end{subfigure}
\hfil
    \begin{subfigure}{0.10\linewidth}
        \includegraphics[width=\linewidth]{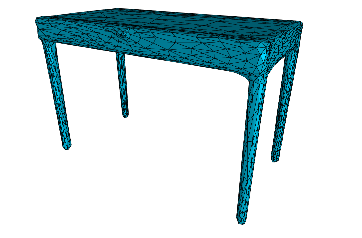}
            \caption*{\tiny$V=1196$ \\ \tiny$F=2388$}
    \end{subfigure}
\hfil
    \begin{subfigure}{0.10\linewidth}
        \includegraphics[width=\linewidth]{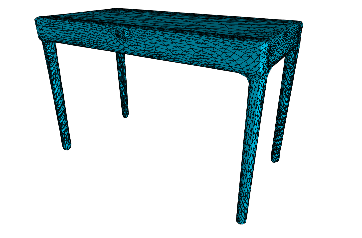}
            \caption*{\tiny$V=3584$ \\ \tiny$F=7164$}
    \end{subfigure}
\hfil
    \begin{subfigure}{0.10\linewidth}
        \includegraphics[width=\linewidth]{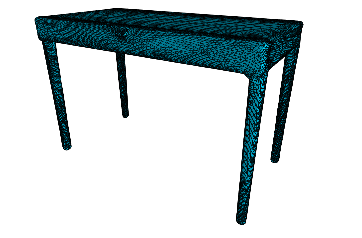}
            \caption*{\tiny$V=10748$ \\ \tiny$F=21492$}
    \end{subfigure}
%%%%%%%%%%%%%%%%%%%%%%%%%%%%%%%%%%%%%%%%%%%%%%%%%%%%%%%%%%%%%%%%%%%%

    \begin{subfigure}{0.10\linewidth}
        \includegraphics[width=\linewidth]{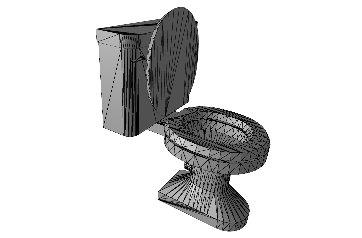}
    \caption*{\tiny$V=4204$ \\ \tiny$F=5364$}
                        \caption*{3D CAD}
    \end{subfigure}
\hfil
    \begin{subfigure}{0.10\linewidth}
        \includegraphics[width=\linewidth]{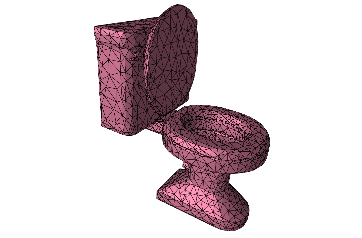}
            \caption*{\tiny$V=2385$ \\ \tiny$F=4770$}
                    \caption*{Cleaned}
    \end{subfigure}
\hfil
    \begin{subfigure}{0.10\linewidth}
        \includegraphics[width=\linewidth]{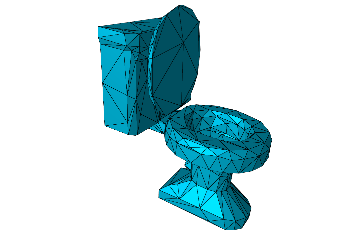}
            \caption*{\tiny$V=400$ \\ \tiny$F=788$}
                    \caption*{Coarse}
    \end{subfigure}
\hfil
    \begin{subfigure}{0.10\linewidth}
        \includegraphics[width=\linewidth]{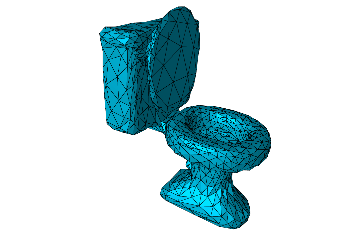}
            \caption*{\tiny$V=1188$ \\ \tiny$F=2364$}
                    \caption*{Sqrt1}
    \end{subfigure}
\hfil
    \begin{subfigure}{0.10\linewidth}
        \includegraphics[width=\linewidth]{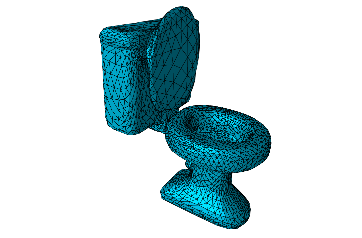}
            \caption*{\tiny$V=3552$ \\ \tiny$F=7092$}
                    \caption*{Sqrt2}
    \end{subfigure}
\hfil
    \begin{subfigure}{0.10\linewidth}
        \includegraphics[width=\linewidth]{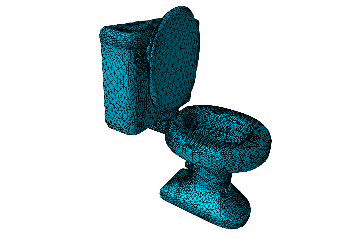}
            \caption*{\tiny$V=10644$ \\ \tiny$F=21276$}
        \caption*{Sqrt3}
    \end{subfigure}
%%%%%%%%%%%%%%%%%%%%%%%%%%%%%%%%%%%%%%%%%%%%%%%%%%%%%%%%%%%%%%%%%%%%
\caption{PolyShape representation. PolyShape processing results on some samples of ModelNet-10 dataset based on $\sqrt{3}$-subdivision. Sqrt1 to Sqrt3 refer to the output of the PolyShape procedure after each level of subdivision. Note that $V$ and $F$ refer to the number of vertices and faces for each shape.}
    \label{fig:SQRT}
\end{figure*}

\renewcommand{\thefigure}{S3}
\begin{figure*}
\captionsetup[subfigure]{justification=centering}
\centering
    \begin{subfigure}{0.10\linewidth}
        \includegraphics[width=\linewidth]{figures/shapes/0/original.png}
            \caption*{\tiny$V=15398$ \\ \tiny$F=19254$}
    \end{subfigure}
\hfil
    \begin{subfigure}{0.10\linewidth}
        \includegraphics[width=\linewidth]{figures/shapes/0/clean.png}
            \caption*{\tiny$V=2500$ \\ \tiny$F=4996$}
    \end{subfigure}
\hfil
    \begin{subfigure}{0.10\linewidth}
        \includegraphics[width=\linewidth]{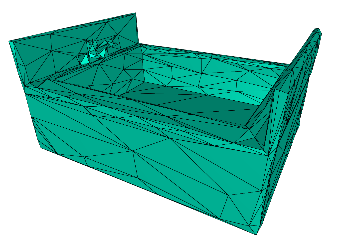}
            \caption*{\tiny$V=400$ \\ \tiny$F=796$}
    \end{subfigure}
\hfil
    \begin{subfigure}{0.10\linewidth}
        \includegraphics[width=\linewidth]{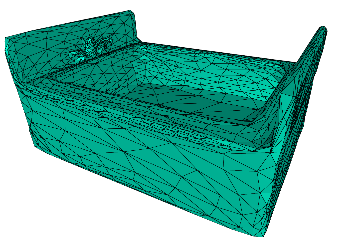}
                    \caption*{\tiny$V=1594$ \\ \tiny$F=3184$}
    \end{subfigure}
\hfil
    \begin{subfigure}{0.10\linewidth}
        \includegraphics[width=\linewidth]{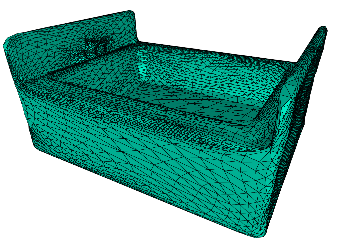}
                    \caption*{\tiny$V=6370$ \\ \tiny$F=12736$}
    \end{subfigure}
\hfil
\begin{subfigure}{0.10\linewidth}
    \includegraphics[width=\linewidth]{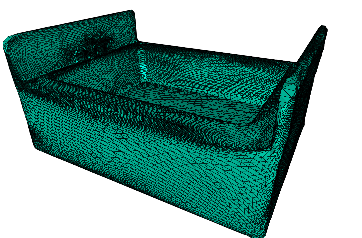}
                \caption*{\tiny$V=25474$ \\ \tiny$F=50944$}
\end{subfigure}
%%%%%%%%%%%%%%%%%%%%%%%%%%%%%%%%%%%%%%%%%%%%%%%%%%%%%%%%%%%%%%%%%%%%

    \begin{subfigure}{0.10\linewidth}
        \includegraphics[width=\linewidth]{figures/shapes/1/original.png}
            \caption*{\tiny$V=584$ \\ \tiny$F=676$}
    \end{subfigure}
\hfil
    \begin{subfigure}{0.10\linewidth}
        \includegraphics[width=\linewidth]{figures/shapes/1/clean.png}
            \caption*{\tiny$V=2502$ \\ \tiny$F=5000$}
    \end{subfigure}
\hfil
    \begin{subfigure}{0.10\linewidth}
        \includegraphics[width=\linewidth]{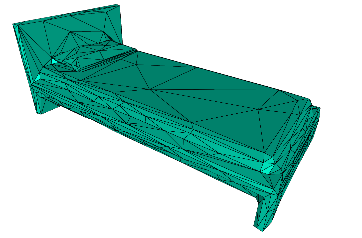}
            \caption*{\tiny$V=400$ \\ \tiny$F=796$}
    \end{subfigure}
\hfil
    \begin{subfigure}{0.10\linewidth}
        \includegraphics[width=\linewidth]{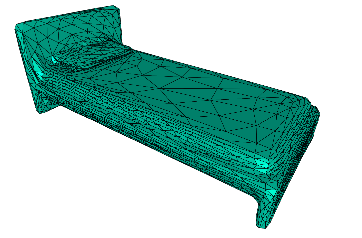}
                    \caption*{\tiny$V=1594$ \\ \tiny$F=3184$}
    \end{subfigure}
\hfil
    \begin{subfigure}{0.10\linewidth}
        \includegraphics[width=\linewidth]{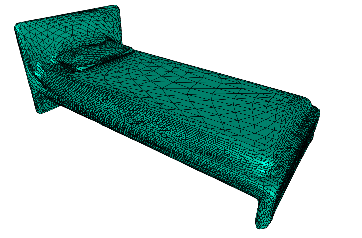}
                    \caption*{\tiny$V=6370$ \\ \tiny$F=12736$}
    \end{subfigure}
\hfil
\begin{subfigure}{0.10\linewidth}
    \includegraphics[width=\linewidth]{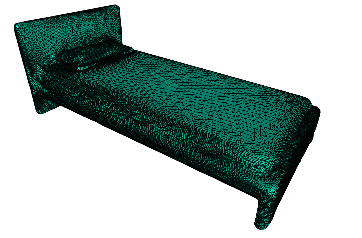}
                \caption*{\tiny$V=25474$ \\ \tiny$F=50944$}
\end{subfigure}
%%%%%%%%%%%%%%%%%%%%%%%%%%%%%%%%%%%%%%%%%%%%%%%%%%%%%%%%%%%%%%%%%%%%

    \begin{subfigure}{0.10\linewidth}
        \includegraphics[width=\linewidth]{figures/shapes/2/original.png}
            \caption*{\tiny$V=584$ \\ \tiny$F=676$}
    \end{subfigure}
\hfil
    \begin{subfigure}{0.10\linewidth}
        \includegraphics[width=\linewidth]{figures/shapes/2/clean.png}
            \caption*{\tiny$V=2502$ \\ \tiny$F=5000$}
    \end{subfigure}
\hfil
    \begin{subfigure}{0.10\linewidth}
        \includegraphics[width=\linewidth]{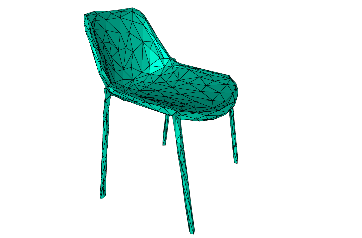}
            \caption*{\tiny$V=400$ \\ \tiny$F=796$}
    \end{subfigure}
\hfil
    \begin{subfigure}{0.10\linewidth}
        \includegraphics[width=\linewidth]{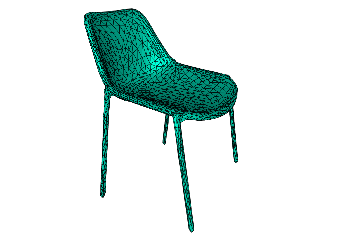}
                    \caption*{\tiny$V=1594$ \\ \tiny$F=3184$}
    \end{subfigure}
\hfil
    \begin{subfigure}{0.10\linewidth}
        \includegraphics[width=\linewidth]{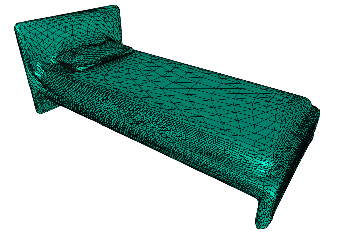}
                    \caption*{\tiny$V=6370$ \\ \tiny$F=12736$}
    \end{subfigure}
\hfil
\begin{subfigure}{0.10\linewidth}
    \includegraphics[width=\linewidth]{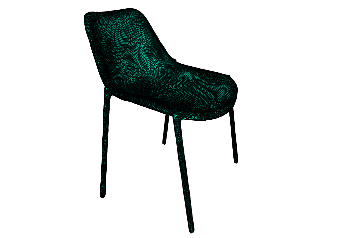}
                \caption*{\tiny$V=25474$ \\ \tiny$F=50944$}
\end{subfigure}
%%%%%%%%%%%%%%%%%%%%%%%%%%%%%%%%%%%%%%%%%%%%%%%%%%%%%%%%%%%%%%%%%%%%

    \begin{subfigure}{0.10\linewidth}
        \includegraphics[width=\linewidth]{figures/shapes/3/original.png}
            \caption*{\tiny$V=1988$ \\ \tiny$F=1376$}
    \end{subfigure}
\hfil
    \begin{subfigure}{0.10\linewidth}
        \includegraphics[width=\linewidth]{figures/shapes/3/clean.png}
            \caption*{\tiny$V=2273$ \\ \tiny$F=4542$}
    \end{subfigure}
\hfil
    \begin{subfigure}{0.10\linewidth}
        \includegraphics[width=\linewidth]{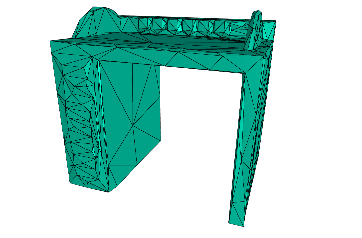}
            \caption*{\tiny$V=400$ \\ \tiny$F=816$}
    \end{subfigure}
\hfil
    \begin{subfigure}{0.10\linewidth}
        \includegraphics[width=\linewidth]{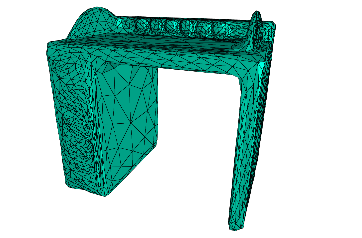}
                    \caption*{\tiny$V=1594$ \\ \tiny$F=3184$}
    \end{subfigure}
\hfil
    \begin{subfigure}{0.10\linewidth}
        \includegraphics[width=\linewidth]{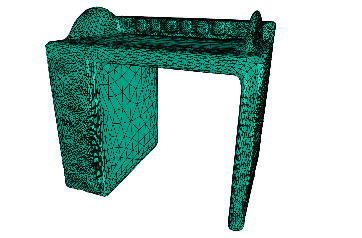}
                    \caption*{\tiny$V=6370$ \\ \tiny$F=12736$}
    \end{subfigure}
\hfil
\begin{subfigure}{0.10\linewidth}
    \includegraphics[width=\linewidth]{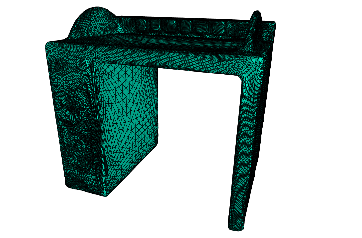}
                \caption*{\tiny$V=25474$ \\ \tiny$F=50944$}
\end{subfigure}
%%%%%%%%%%%%%%%%%%%%%%%%%%%%%%%%%%%%%%%%%%%%%%%%%%%%%%%%%%%%%%%%%%%%

    \begin{subfigure}{0.10\linewidth}
        \includegraphics[width=\linewidth]{figures/shapes/4/original.png}
                    \caption*{\tiny$V=3958$ \\  \tiny$F=3132$}
    \end{subfigure}
\hfil
    \begin{subfigure}{0.10\linewidth}
        \includegraphics[width=\linewidth]{figures/shapes/4/clean.png}
            \caption*{\tiny$V=2502$ \\ \tiny$F=5000$}
    \end{subfigure}
\hfil
    \begin{subfigure}{0.10\linewidth}
        \includegraphics[width=\linewidth]{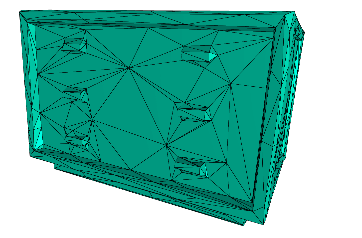}
            \caption*{\tiny$V=400$ \\ \tiny$F=796$}
    \end{subfigure}
\hfil
    \begin{subfigure}{0.10\linewidth}
        \includegraphics[width=\linewidth]{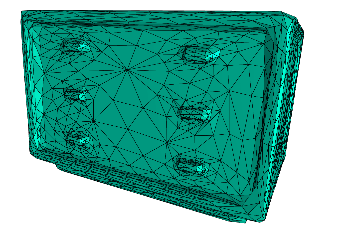}
                    \caption*{\tiny$V=1594$ \\ \tiny$F=3184$}
    \end{subfigure}
\hfil
    \begin{subfigure}{0.10\linewidth}
        \includegraphics[width=\linewidth]{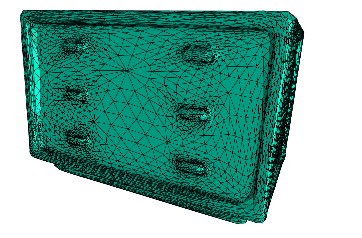}
                    \caption*{\tiny$V=6370$ \\ \tiny$F=12736$}
    \end{subfigure}
\hfil
\begin{subfigure}{0.10\linewidth}
    \includegraphics[width=\linewidth]{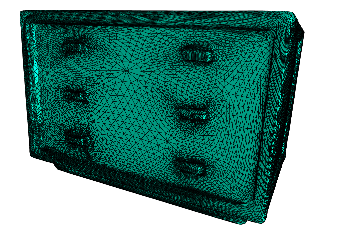}
                \caption*{\tiny$V=25474$ \\ \tiny$F=50944$}
\end{subfigure}
%%%%%%%%%%%%%%%%%%%%%%%%%%%%%%%%%%%%%%%%%%%%%%%%%%%%%%%%%%%%%%%%%%%%

    \begin{subfigure}{0.10\linewidth}
        \includegraphics[width=\linewidth]{figures/shapes/5/original.png}
                    \caption*{\tiny$V=2482$ \\ \tiny$F=1794$}
    \end{subfigure}
\hfil
    \begin{subfigure}{0.10\linewidth}
        \includegraphics[width=\linewidth]{figures/shapes/5/clean.png}
                    \caption*{\tiny$V=2500$ \\ \tiny$F=5000$}
    \end{subfigure}
\hfil
    \begin{subfigure}{0.10\linewidth}
        \includegraphics[width=\linewidth]{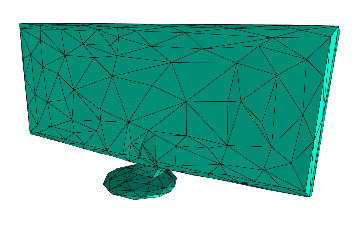}
                    \caption*{\tiny$V=400$ \\ \tiny$F=800$}
    \end{subfigure}
\hfil
    \begin{subfigure}{0.10\linewidth}
        \includegraphics[width=\linewidth]{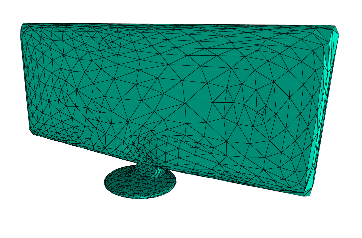}
                \caption*{\tiny$V=1600$ \\ \tiny$F=3200$}
    \end{subfigure}
\hfil
    \begin{subfigure}{0.10\linewidth}
        \includegraphics[width=\linewidth]{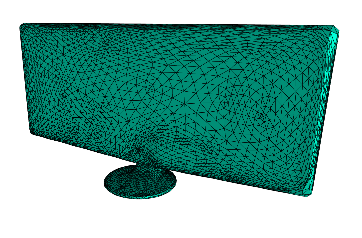}
                        \caption*{\tiny$V=6400$ \\ \tiny$F=12800$}
    \end{subfigure}
\hfil
\begin{subfigure}{0.10\linewidth}
    \includegraphics[width=\linewidth]{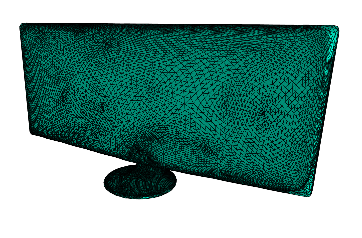}
                        \caption*{\tiny$V=25600$ \\ \tiny$F=51200$}
\end{subfigure}
%%%%%%%%%%%%%%%%%%%%%%%%%%%%%%%%%%%%%%%%%%%%%%%%%%%%%%%%%%%%%%%%%%%%

    \begin{subfigure}{0.10\linewidth}
        \includegraphics[width=\linewidth]{figures/shapes/6/original.png}
                    \caption*{\tiny$V=41528$ \\ \tiny$F=22000$}
    \end{subfigure}
\hfil
    \begin{subfigure}{0.10\linewidth}
        \includegraphics[width=\linewidth]{figures/shapes/6/clean.png}
                    \caption*{\tiny$V=2037$ \\ \tiny$F=4074$}
    \end{subfigure}
\hfil
    \begin{subfigure}{0.10\linewidth}
        \includegraphics[width=\linewidth]{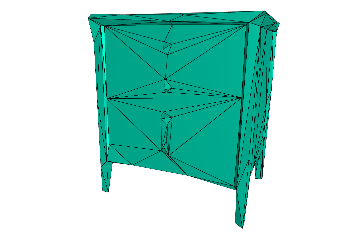}
                    \caption*{\tiny$V=400$ \\ \tiny$F=942$}
    \end{subfigure}
\hfil
    \begin{subfigure}{0.10\linewidth}
        \includegraphics[width=\linewidth]{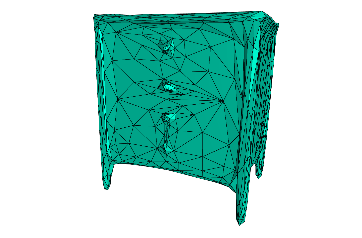}
             \caption*{\tiny$V=1600$ \\ \tiny$F=3200$}
    \end{subfigure}
\hfil
    \begin{subfigure}{0.10\linewidth}
        \includegraphics[width=\linewidth]{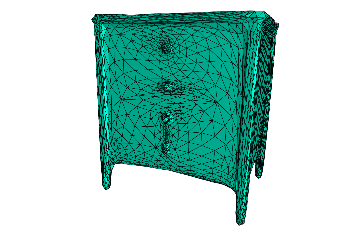}
                \caption*{\tiny$V=6400$ \\ \tiny$F=12800$}
    \end{subfigure}
\hfil
\begin{subfigure}{0.10\linewidth}
    \includegraphics[width=\linewidth]{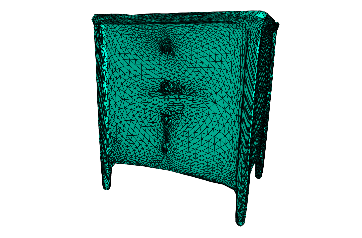}
             \caption*{\tiny$V=25600$ \\ \tiny$F=51200$}
\end{subfigure}
%%%%%%%%%%%%%%%%%%%%%%%%%%%%%%%%%%%%%%%%%%%%%%%%%%%%%%%%%%%%%%%%%%%%

    \begin{subfigure}{0.10\linewidth}
        \includegraphics[width=\linewidth]{figures/shapes/7/original.png}
                    \caption*{\tiny$V=7808$ \\ \tiny$F=13980$}
    \end{subfigure}
\hfil
    \begin{subfigure}{0.10\linewidth}
        \includegraphics[width=\linewidth]{figures/shapes/7/clean.png}
                    \caption*{\tiny$V=2496$ \\ \tiny$F=4988$}
    \end{subfigure}
\hfil
    \begin{subfigure}{0.10\linewidth}
        \includegraphics[width=\linewidth]{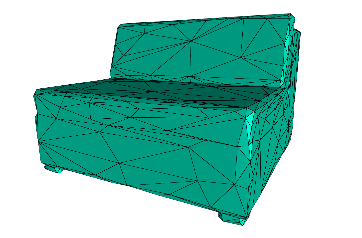}
                    \caption*{\tiny$V=400$ \\ \tiny$F=798$}
    \end{subfigure}
\hfil
    \begin{subfigure}{0.10\linewidth}
        \includegraphics[width=\linewidth]{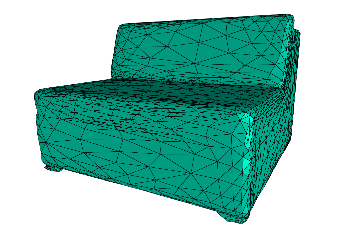}
                            \caption*{\tiny$V=1594$ \\ \tiny$F=3184$}
    \end{subfigure}
\hfil
    \begin{subfigure}{0.10\linewidth}
        \includegraphics[width=\linewidth]{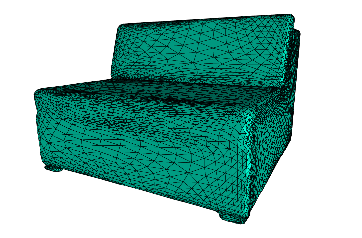}
                            \caption*{\tiny$V=6370$ \\ \tiny$F=12736$}
    \end{subfigure}
\hfil
\begin{subfigure}{0.10\linewidth}
    \includegraphics[width=\linewidth]{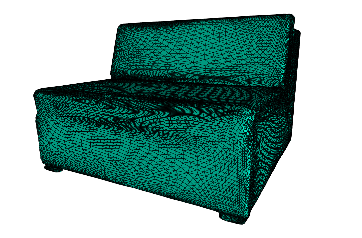}
                        \caption*{\tiny$V=25474$ \\ \tiny$F=50944$}
\end{subfigure}
%%%%%%%%%%%%%%%%%%%%%%%%%%%%%%%%%%%%%%%%%%%%%%%%%%%%%%%%%%%%%%%%%%%%

    \begin{subfigure}{0.10\linewidth}
        \includegraphics[width=\linewidth]{figures/shapes/8/original.png}
                    \caption*{\tiny$V=1074$ \\ \tiny$F=694$}
    \end{subfigure}
\hfil
    \begin{subfigure}{0.10\linewidth}
        \includegraphics[width=\linewidth]{figures/shapes/8/clean.png}
                    \caption*{\tiny$V=2502$ \\ \tiny$F=5000$}
    \end{subfigure}
\hfil
    \begin{subfigure}{0.10\linewidth}
        \includegraphics[width=\linewidth]{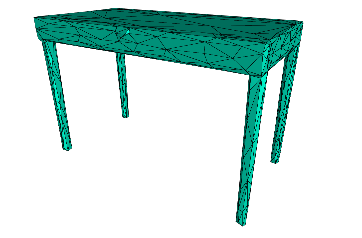}
                    \caption*{\tiny$V=400$ \\ \tiny$F=796$}
    \end{subfigure}
\hfil
    \begin{subfigure}{0.10\linewidth}
        \includegraphics[width=\linewidth]{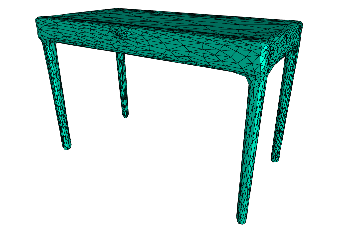}
                            \caption*{\tiny$V=1594$ \\ \tiny$F=3184$}
    \end{subfigure}
\hfil
    \begin{subfigure}{0.10\linewidth}
        \includegraphics[width=\linewidth]{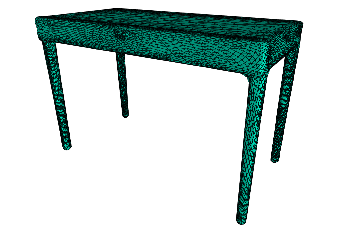}
                            \caption*{\tiny$V=6370$ \\ \tiny$F=12736$}
    \end{subfigure}
\hfil
\begin{subfigure}{0.10\linewidth}
    \includegraphics[width=\linewidth]{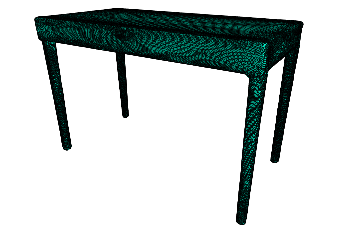}
                        \caption*{\tiny$V=25474$ \\ \tiny$F=50944$}
\end{subfigure}
%%%%%%%%%%%%%%%%%%%%%%%%%%%%%%%%%%%%%%%%%%%%%%%%%%%%%%%%%%%%%%%%%%%%

    \begin{subfigure}{0.10\linewidth}
        \includegraphics[width=\linewidth]{figures/shapes/9/original.png}    \caption*{\tiny$V=4204$ \\ \tiny$F=5364$}
                             \caption*{3D CAD}
    \end{subfigure}
\hfil
    \begin{subfigure}{0.10\linewidth}
        \includegraphics[width=\linewidth]{figures/shapes/9/clean.png}
                    \caption*{\tiny$V=2385$ \\ \tiny$F=4770$}
                                         \caption*{Cleaned}
    \end{subfigure}
\hfil
    \begin{subfigure}{0.10\linewidth}
        \includegraphics[width=\linewidth]{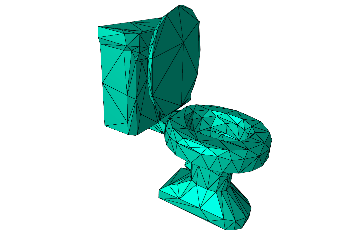}
                    \caption*{\tiny$V=400$ \\ \tiny$F=788$}
                     \caption*{Coarse}
    \end{subfigure}
\hfil
    \begin{subfigure}{0.10\linewidth}
        \includegraphics[width=\linewidth]{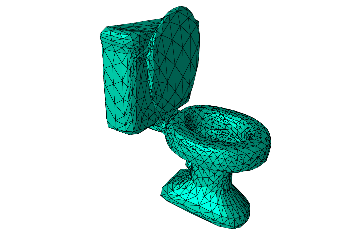}
                            \caption*{\tiny$V=1600$ \\ \tiny$F=3200$}
                             \caption*{PTQ1}
    \end{subfigure}
\hfil
    \begin{subfigure}{0.10\linewidth}
        \includegraphics[width=\linewidth]{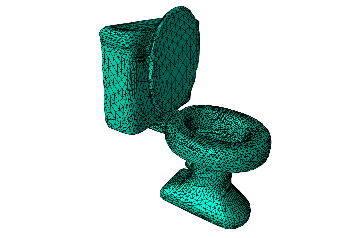}
                            \caption*{\tiny$V=6400$ \\ \tiny$F=12800$}
                             \caption*{PTQ2}
    \end{subfigure}
\hfil
\begin{subfigure}{0.10\linewidth}
    \includegraphics[width=\linewidth]{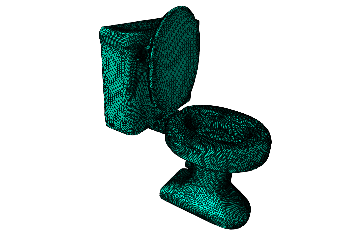}
                        \caption*{\tiny$V=25600$ \\ \tiny$F=51200$}
                        \caption*{PTQ3}
\end{subfigure}    
\caption{PolyShape representation. PolyShape processing results on some samples of ModelNet-10 dataset based on PTQ. PTQ1 to PTQ3 refer to the output of the PolyShape procedure after each level of subdivision. Note that $V$ and $F$ refer to the number of vertices and faces for each shape.}
    \label{fig:PTQ}
    \end{figure*}

\end{appendices}

\end{document}